\documentclass{bmvc2k}

\usepackage{times}
\usepackage{epsfig}
\usepackage{graphicx}
\usepackage{amsmath}
\usepackage{amssymb}

\usepackage[bigfiles]{pdfbase}
\ExplSyntaxOn
\cs_new:Npn\embedvideo#1#2{
  \leavevmode
  \pbs_pdfobj:nnn{}{fstream}{{}{#2}}
  \pbs_pdfobj:nnn{}{dict}{
    /Type/Filespec/F~(#2)/UF~(#2)
    /EF~<</F~\pbs_pdflastobj:>>
  }
  \tl_set:Nx\video{\pbs_pdflastobj:}%
  \pbs_pdfobj:nnn{}{dict}{
    /Type/RichMediaInstance/Subtype/Video
    /Asset~\video
    /Params~<</Binding/Foreground>>
  }
  \pbs_pdfobj:nnn{}{dict}{
    /Type/RichMediaConfiguration/Subtype/Video
    /Instances~[\pbs_pdflastobj:]
  }
  \pbs_pdfobj:nnn{}{dict}{
    /Type/RichMediaContent
    /Assets~<<
      /Names~[(#2)~\video]
    >>
    /Configurations~[\pbs_pdflastobj:]
  }
  \tl_set:Nx\rmcontent{\pbs_pdflastobj:}%
  \pbs_pdfobj:nnn{}{dict}{
    /Activation~<<
      /Condition/XA
      /Presentation~<<
        /Transparent~true
        /Style/Embedded
        /PassContextClick~true
      >>
    >>
    /Deactivation~<</Condition/PC>>
  }
  \hbox_set:Nn\l_tmpa_box{#1}
  \tl_set:Nx\l_box_wd_tl{\dim_use:N\box_wd:N\l_tmpa_box}
  \tl_set:Nx\l_box_ht_tl{\dim_use:N\box_ht:N\l_tmpa_box}
  \tl_set:Nx\l_box_dp_tl{\dim_use:N\box_dp:N\l_tmpa_box}
  \pbs_pdfxform:nnnnn{1}{1}{}{}{\l_tmpa_box}
  \pbs_pdfannot:nnnn{\l_box_wd_tl}{\l_box_ht_tl}{\l_box_dp_tl}{
    /Subtype/RichMedia
    /BS~<</W~0/S/S>>
    /Contents~(embedded~video~file:#2)
    /NM~(rma:#2)
    /AP~<</N~\pbs_pdflastxform:>>
    /RichMediaSettings~\pbs_pdflastobj:
    /RichMediaContent~\rmcontent
  }
  \phantom{#1}
}%
\ExplSyntaxOff

\usepackage{fancyref}
\usepackage{xcolor}
\usepackage{booktabs}
\usepackage{textcomp}
\usepackage{multirow}
\usepackage{multicol}
\usepackage{bm}
\usepackage{multimedia}
\definecolor{captioncolor}{rgb}{0,0,.4}
\usepackage[font={color=captioncolor},skip=2pt]{caption}
\usepackage{float}

\usepackage{xspace}

\title{Learning to Deblur and Rotate Motion-Blurred Faces}

\addauthor{Givi Meishvili}{givi.meishvili@inf.unibe.ch}{1}
\addauthor{Attila Szab\'{o}}{attila.szabo@huawei.com}{3}
\addauthor{Simon Jenni}{jenni@adobe.com}{2}
\addauthor{Paolo Favaro}{paolo.favaro@inf.unibe.ch}{1}

\addinstitution{
University of Bern,
Switzerland\\
}
\addinstitution{
Adobe Research\\
}
\addinstitution{
Huawei, Noah's Ark Lab\\
\tiny{(work was done before joining)}
}

\runninghead{Meishvili et al.}{Learning to Deblur and Rotate Motion-Blurred Faces}

\def\eg{\emph{e.g}\bmvaOneDot}

\def\etal{\emph{et al}\bmvaOneDot}
\def\ie{\emph{i.e}\bmvaOneDot}

\usepackage{media9}

\begin{document}

\maketitle
\begin{abstract}
We propose a solution to the novel task of rendering sharp videos from new viewpoints from a single motion-blurred image of a face. 
Our method\footnote{Code and data available at \href{https://gmeishvili.github.io/deblur_and_rotate_motion_blurred_faces/index.html}{https://gmeishvili.github.io/deblur\_and\_rotate\_motion\_blurred\_faces/index.html}.} handles the complexity of face blur by implicitly learning the geometry and motion of faces through the joint training on three large datasets: FFHQ and 300VW, which are publicly available, and a new Bern Multi-View Face Dataset (BMFD) that we built. 
The first two datasets provide a large variety of faces and allow our model to generalize better. 
BMFD instead allows us to introduce multi-view constraints, which are crucial to synthesizing sharp videos from a new camera view. 
It consists of high frame rate synchronized videos from multiple views of several subjects displaying a wide range of facial expressions. 
We use the high frame rate videos to simulate realistic motion blur through averaging. 
Thanks to this dataset, we train a neural network to reconstruct a 3D video representation from a single image and the corresponding face gaze. 
We then provide a camera viewpoint relative to the estimated gaze and the blurry image as input to an encoder-decoder network to generate a video of sharp frames with a novel camera viewpoint.
We demonstrate our approach on test subjects of our multi-view dataset and VIDTIMIT.
\end{abstract}
\section{Introduction}
\begin{figure*}[t]
\centering
\includegraphics[width=0.8\textwidth]{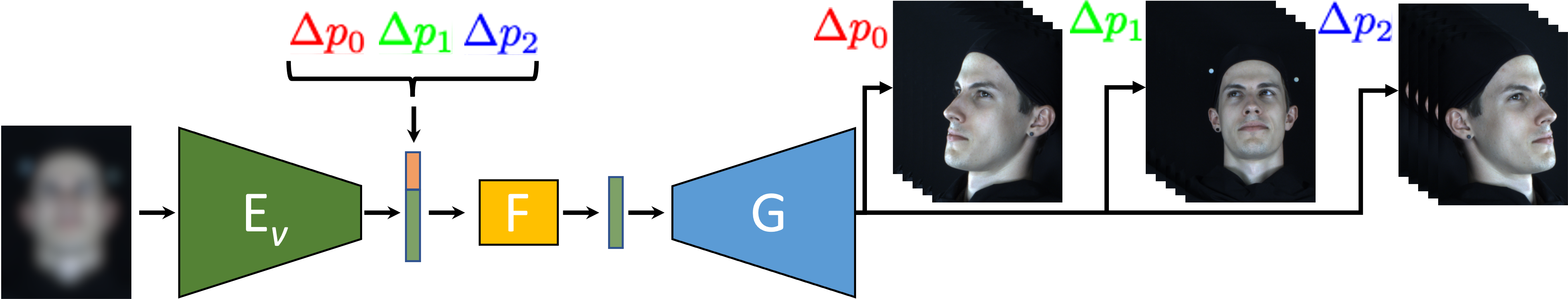}
\caption{
\textbf{Overview of our system during inference.}
The encoder $E_v$ encodes a blurry image into a sequence of latent codes that are then manipulated based on a relative viewpoint (\eg, $\Delta p_{0}, \Delta p_{1},$ or $\Delta p_{2}$) via the fusion network $F$ to produce encodings of images from a novel view. Finally, the generator $G$ maps the novel view encodings to the image space.
 \label{fig:inference_overview}}
\end{figure*}
Faces are a fundamental subject in image processing and in recognition due to their role in applications such as teleconferencing, video surveillance, biometrics, video analytics, entertainment, and smart shopping, just to name a few.
In particular, in the case of teleconferencing, the interaction is found to be more engaging when the person on the screen looks towards the receiver \cite{gaze_importance}. However, to achieve this configuration it is necessary to look directly into the camera. Unfortunately, this does not allow to watch the person on the screen that one talks to. A solution to this issue is to design a system that can render the captured face from an arbitrary viewpoint. Then, it becomes possible to dynamically adapt the gaze of the face on the screen to ensure that it aims at the observer. Moreover, because of the low frame rate of web cameras, especially when used in low light, it becomes important to solve the above task in the presence of motion blur. Since a blurry image is the result of averaging several sharp frames \cite{Nah_2017_CVPR}, one could pose the problem of recovering not one, but a sequence of sharp frames from the single blurry input. This capability enables a smooth temporal rendering of the video. In addition, one might use this capability to deal with a limited connection bandwidth. Current software fits the available bandwidth by reducing the frame rate of the captured video. However, instead of selecting temporally distant frames, one could also transmit the average of several frames and then restore the original (high) frame rate at the destination terminal.
In this paper, we present a method that recovers a sharp video rendered from an arbitrary viewpoint from a single blurry image of a face. Figure~\ref{fig:inference_overview} shows our model during the inference stage. We design a neural network and a training scheme to remove motion blur from an image and produce a video of sharp frames with a general viewpoint. 
Our neural network is built in two steps: First, by training a generative model that outputs face images from zero-mean Gaussian noise, which we call the \emph{latent space}, and then by training encoders to map images to the latent space. 
The primary motivation to use a generative model is that face rotations can be handled more easily in the latent space than in image space. This property was recently observed for generative adversarial networks \cite{voynov2020unsupervised}.
One encoder is trained so that, when concatenated with the generator, it autoencodes face images. 
Then, rather than using sharp images as targets in a loss, we use their encodings, the latent vectors, as targets. 
As a second step, we obtain a blurry image by averaging several sharp frames. Then, we train a second encoder to map the blurry image to a sequence of latent vectors that match the target latent vectors corresponding to the original sharp frames. 
Finally, the change of the face viewpoint requires the availability of the latent vectors corresponding to the same face instance, but rotated. To the best of our knowledge, there are no public face datasets with such data. Thus, we built a novel multiview face dataset. This dataset consists of videos captured at 112 fps of 52 individuals performing several expressions. Thanks to the high frame rate, we can simulate realistic blur through temporal averaging. Each performance is captured simultaneously from 8 different viewpoints so that it is possible to encode multiple views of the same temporal instance into target latent vectors and then train a fusion network to map the latent vector of one view and a relative viewpoint to the latent vector of another view of the same face instance. The relative viewpoint we provide as input should be the relative pose between the input and the output face poses. While we can use the viewpoint information from our calibrated camera rig during training, with new data, this information may be unknown. Hence, we also train a neural network to estimate the head pose. The network learns to map an image to Basel Face Model \cite{bfm09} (BFM) parameters, such that, when rendered (through a differential renderer), it matches the input image. 

\noindent \textbf{Contributions.}
We make the following contributions: \textit{(i)} We introduce BMFD, a novel high frame rate multi-view face dataset that allows more accurate modeling of natural motion blur and the incorporation of 3D constraints; \textit{(ii)} As a novel task enabled through this data, we propose a model that, given a blurry face image, can synthesize a sharp video from arbitrary views; \textit{(iii)} We demonstrate this capability on our multiview dataset and VIDTIMIT \cite{VIDTIMIT}. 

\section{Prior Work}
\noindent \textbf{3D Face Reconstruction.}
3D morphable models (3DMM) \cite{Blanz_1999_3DMM} provide an interpretable generative model of faces in the form of a linear combination of base shapes.
In the past decades many improvements were made using more data, better scanning devices or more detailed modelling \cite{bfm09, Booth_2016_CVPR, Bolkart_2016_CVPR, Peng_2017_CVPR, Tran_2018_CVPR_1,  Ranjan_2018_ECCV, Liu_2019_ICCV, Ploumpis_2019_CVPR, Tran_2019_CVPR, egger20203d, Yang_2020_CVPR}.
3D face reconstruction can be cast as regressing the parameters of such 3DMMs. 
The model parameters can be fit using multi-view images \cite{Piotraschke_2016_CVPR, Wu_2019_CVPR, Sanyal_2019_CVPR,Tewari_2019_CVPR,Roth_2016_CVPR}.
Since 3DMMs provide a strong shape prior, they also enable single-image 3D reconstruction \cite{Tewari_2018_CVPR, Booth_2017_CVPR, Kim_2018_CVPR}. These methods learn to estimate the model parameters by matching input images with differentiable rendering techniques \cite{Genova_2018_CVPR, kato2018neural, szabo2019unsupervised, Zhu_2020_CVPR}.
We also leverage a 3DMM to learn a controllable representation of faces. In our work, these representations are used to manipulate the latent space of a StyleGAN generator.

\noindent \textbf{Face Deblurring.} While we focus on modern learning-based approaches, there exist specialized classic approaches such as \cite{Pan_ECCV_2014_face_deblur}. Several works designed specialized neural network architectures to target face deblurring. \cite{Chrysos_2017_CVPR_Workshops} performed face alignment to the input of the network. 
\cite{chrysos_face_deblur} introduced a two-stage architecture where the first stage restores low frequency and the second stage restores high-frequency content.
\cite{Jin_2018_CVPR_Workshops} designed computationally efficient architecture that exploits a very large receptive field.

Some methods incorporate additional information in the form of semantic label maps  \cite{Shen_2018_CVPR,ryasarla_UMSN} or 3D priors from a 3DMM \cite{Ren_2019_ICCV}.
\cite{Lu_2019_CVPR} disentangled image content and blur and exploit cycle-consistency to learn deblurring in the unsupervised, \ie, unpaired setting. 
Face deblurring has also been combined with super-resolution by restoring high-resolution facial images from blurry low-resolution images \cite{Xu_2017_ICCV_1,SONG_IJCV_2019_FHD}.
Our approach is different in that we invert a state-of-the-art generative model of face images.

Several works in the field focused on extracting a sharp video from a single motion blurred image \cite{Jin_2018_CVPR,Purohit_2019_CVPR}. \citet{Jin_2019_CVPR} introduced the task and solution of generating a sharp slow-motion video given a low frame rate blurry video.

\noindent \textbf{Novel Face View Synthesis.}
Since our method allows the rendering of deblurred faces from novel views, we briefly discuss relevant work on novel face view synthesis. 
\cite{xu2019view} use an encoder-decoder architecture. The encoder extracts view independent features, which are fed to the decoder along with sampled camera parameters. Realism and pose consistency are enforced via GANs.
\cite{hu2018pose} use face landmarks to guide and condition the novel face view reconstruction. 
A special case of novel-view synthesis on faces is face frontalization \cite{masi2018deep,hassner2015effective,zhang2018face}.  \cite{huang2017beyond} design a GAN architecture for face frontalization. Their generator consists of two pathways: A global pathway processes the whole image, and a local pathway processes local patches extracted at landmarks. Tackling the opposite problem, \cite{zhao2017dual} train a GAN to generate silhouette images in order to reduce the pose bias in existing face datasets. To the best of our knowledge, we are the first to deblur and synthesize frames from a novel view simultaneously. 

\begin{figure*}[!ht]
\centering
\includegraphics[width=0.96\textwidth]{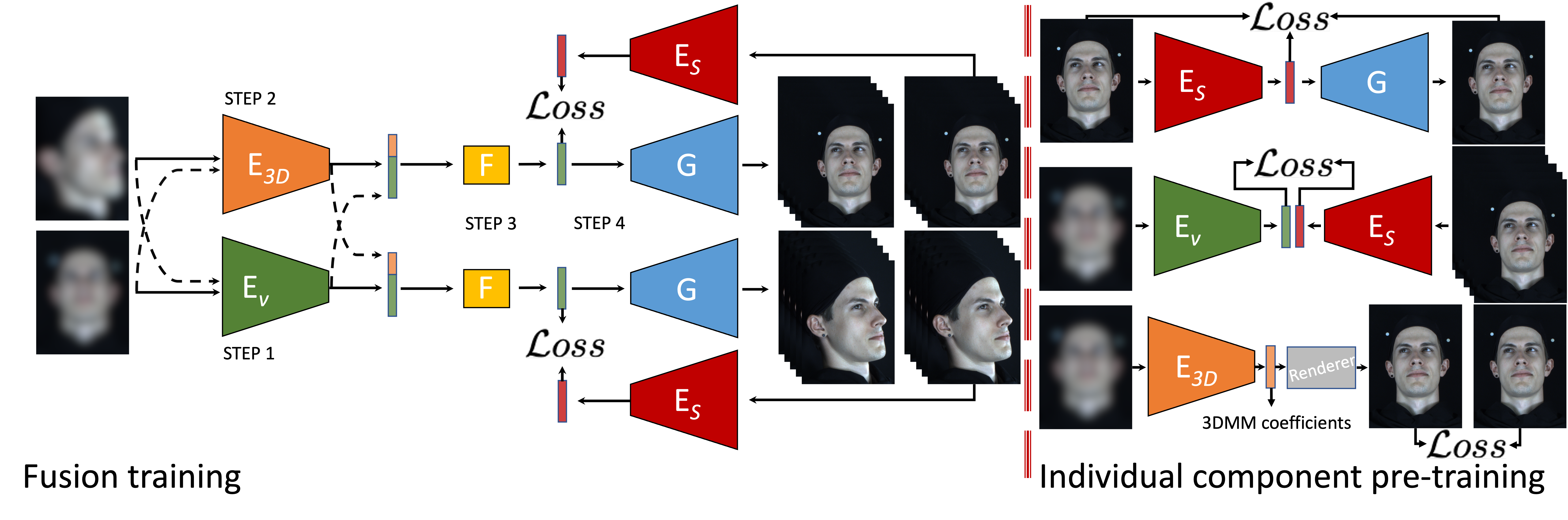}
\caption{\textbf{Overview of the model architecture.} 
From top to bottom, on the right side of the figure, we show the individual pre-training stages of encoders: $E_s$, $E_v$, and $E_{3D}$. A sharp image generator $G$, is pre-trained using StyleGAN2.
The training of the model $F$ is shown on the right side of the figure. The encoder $E_v$ encodes a blurry image into a sequence of latent codes corresponding to a sequence of sharp frames (step 1). Pose information is extracted via the viewpoint encoder $E_{3D}$, which is trained to regress the coefficients of a 3DMM (step 2). The predicted sharp latent codes are then manipulated based on the pose encodings via the fusion network $F$ to produce latent codes of images from a novel view (step 3). Finally, generator $G$ maps the novel view encodings to the image space (step 4).
\label{fig:model}}
\end{figure*}

\section{Model}
Our goal is to design a model that can generate a sharp video of a face from a single motion-blurred image.
Additionally, we want to synthesize novel views of these videos, \ie, rotate the reconstructions.
We design a modular architecture to achieve this goal (see Figure~\ref{fig:model}). We give an overview of the components here and provide more details in the following subsections. 
The bedrock of our approach is a generative model $G$ of sharp face images. 
We describe how we can leverage the generative model $G$ by learning an inverse mapping $E_{s}$ from image-space to $G$'s latent space in section \ref{sec:gan_training}. 
The sharp image encoder $E_{s}$ then acts as a teacher for a blurry image encoder $E_{v}$.
In section \ref{sec:blurry2MultiZ_training} we describe how to train $E_{v}$ to predict latent codes of multiple sharp frames by using encodings of $E_{s}$ as targets. 
To perform novel view synthesis, we require to capture the 3D viewpoint of the face. 
To this end, we learn a viewpoint extractor $E_{3D}$ that maps a blurry image to coefficients of a 3DMM. We describe how to train $E_{3D}$ using a differentiable renderer in section~\ref{sec:3dvideo_training}.
The viewpoint from $E_{3D}$ can then be used to manipulate the latent codes of a blurry image obtained through $E_{v}$. We do so by training a model $F$ that, given relative viewpoint changes obtained through $E_{3D}$ and latent codes from $E_{v}$, outputs updated latent codes corresponding to the desired change of viewpoint. This process is described in section~\ref{sec:fusion_training}. 
\begin{figure*}[t]
\centering
\includegraphics[width=1.0\textwidth]{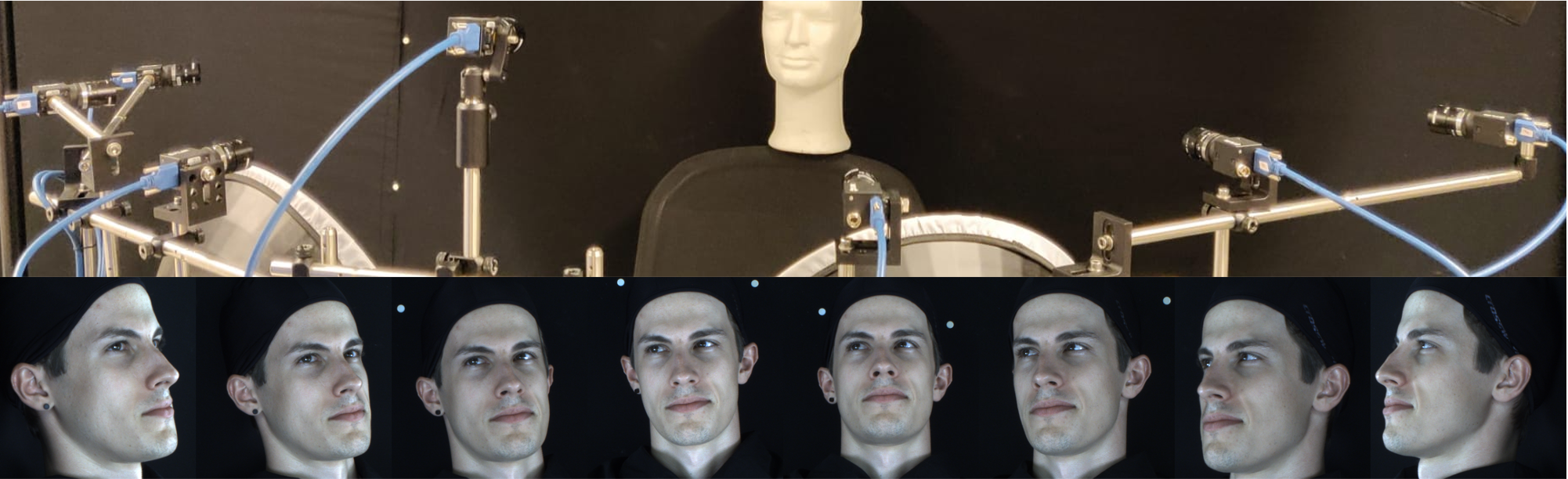}
\caption{
\textbf{Overview of our multi-view video capture setup.}
In our lab setting, we arranged eight high-speed cameras in a circular grid. The cameras capture synchronized videos of participants performing a wide range of facial expressions from a wide variety of viewpoints. We show an example of 8 synchronized views of one of the 52 participants in BMFD. Background and clothing are black, allowing the easier extraction of skin regions. 
 \label{fig:3Dfaces}}
\end{figure*}

\noindent \textbf{Data.}
Our dataset consists of a set of sharp frames $\{y_i\}_{i=1}^N$. We synthesize blurry images by averaging $2m+1$ consecutive frames, \ie, $x_i=\frac{1}{2m+1} \sum_{j=i-m}^{i+m}y_i$. As targets we define a sequence of 5 sharp frames  $\bm{y}_i=[y_{i-m}, y_{i-m/2}, y_{i}, y_{i+m/2}, y_{i+m}]$. 
The training dataset then is given by $\mathcal{D} = \big\{ (x_i^\nu, \bm{y}_i^\nu)\mid i=1, \ldots,n; \nu=1,\ldots,8 \big\}$, where the superscript $\nu$ indicates the viewpoint (we omit $\nu$ when it is not needed). 

\subsection{Bern Multi-View Face Dataset}
\label{sec:dataset}
Most prior face deblurring methods tackle the shift-invariant blur case, \ie, blur that might arise from camera shake. 
Training data for such methods can be synthesized by convolving sharp face images with random blur kernels \cite{Jin_2018_CVPR_Workshops,Lu_2019_CVPR,Shen_2018_CVPR}.
However, such models do not generalize well to blur caused by face motion since the resulting blurs are no longer spatially invariant.
To tackle motion blur, Ren~\etal~\cite{Ren_2019_ICCV} generate training data by averaging consecutive frames of the 300-VW dataset \cite{Shen_2015_ICCV_Workshops}.
This is a valid approximation of natural motion blur when the frame rate of the videos is sufficiently high. 
Since the 300-VW data has a relatively low frame rate of 25-30 fps, the resulting synthetic motion blurs are not always of high quality and can exhibit ghosting artifacts.
Additionally, existing face datasets exhibit a pose bias, with most images showing faces in a frontal pose. Methods trained on such data can show poor generalization to non-frontal views. 

To overcome these limitations, we introduce a dataset of high-speed, multi-view face videos. 
The faces of 52 participants were captured in a lab setting from 8 fixed viewpoints simultaneously. 
The cameras were arranged in a circular grid, ensuring that the faces are captured from all sides (see Figure~\ref{fig:3Dfaces}). 
Videos are captured at 112 frames per second at a resolution of 1440$\times$1080. The duration of the recordings ranges between 75 and 90 seconds. 

\subsection{Inverting a Generative Face Model}
\label{sec:gan_training}
In order to generate novel views of a video sequence, we rely on a generative model of face images with a latent space where manipulations that change viewpoints are feasible. 
Consequently, we chose to train a SyleGAN2 \cite{Karras2019stylegan2} as the generator $G$ of sharp face images. 
SyleGAN2 provides state-of-the-art image quality and a smooth, disentangled latent space.
To reconstruct or manipulate a given face image $y_i$, we require a corresponding latent code $z_i$, s.t. $G(z_i)=y_i$. 
To this end, we train a sharp image encoder $E_s$ to invert the generator $G$, \ie, we want that $G(E_s(y))=y$.
We adopt the inversion strategy of Meishvili~\etal~\cite{Meishvili_2020_CVPR}, where the encoder $E_{s}$ is trained while the generator $G$ is fine-tuned. The training objective is given by
\begin{equation}
\label{eq:fixed_generator}
	\min_{E_{s},G} \sum_{i=1}^n \ell_{s} \left( G\left(E_{s}(y_{i})\right), y_i \right)  + \lambda_g\big|G_\text{init} - G\big|_2^2+ \lambda_s \big|1-|E_{s}(y_{i})|\big|,
\end{equation}
\noindent where $\ell_{s}$ represents the following combination of different reconstruction losses:
\begin{equation}
\ell_{s} \left( x, y \right) = \lambda_{id}\mathcal{L}_{id}\left( x, y \right) + \lambda_{per}\mathcal{L}_{per}\left( x, y \right) + \lambda_{edge}\mathcal{L}_{edge}\left( x, y \right) + \big|x-y\big|,
\nonumber
\end{equation} 
\begin{equation}
\resizebox{.99 \textwidth}{!} 
{
$\mathcal{L}_{id}\left( x, y \right)=1-\frac{<\phi_{id}(x), \phi_{id}\left(y\right)>}{|\phi_{id}(x)| \cdot\left|\phi_{id}\left(y\right)\right|}, \quad \mathcal{L}_{per}\left( x, y \right)=\big|\phi_{per}(x)-\phi_{per}(y)\big|_2^2, \quad \mathcal{L}_{edge}\left( x, y \right)=\big|\texttt{S}(x)-\texttt{S}(y)\big|.$}
\nonumber
\end{equation}
$\mathcal{L}_{id}$ is a term minimizing the cosine between embeddings of a pre-trained identity classification network $\phi_{id}$ of \citet{VGGFACE2}.
$\mathcal{L}_{per}$ is a perceptual loss on features of an ImageNet pre-trained VGG16 network $\phi_{per}$ \cite{vgg2014}.
$\mathcal{L}_{edge}$ is a Sobel edge matching term.
We used a naive Bayes classifier with Gaussian Mixture Models trained on a skin image dataset from \cite{skin_mask_ref} to double the contribution of the skin pixels in all the losses. 

$\lambda_g=1$ controls how much $G$ is allowed to deviate from the initial generator parameters $G_\text{init}$ (before fine-tuning), and $\lambda_s=1$ softly enforces that the predicted latent codes lie on the unit hypersphere. 
During training, we gradually relax $\lambda_g$ until we reach the desired reconstruction quality.
Similar to \cite{Meishvili_2020_CVPR} we regress multiple latent codes per frame, each injected at different layers of the StyleGAN2. Thus $E_s(y_i)=z_i\in \mathbb{R}^{14\times512}$. Weights controlling the contribution of each term are set as follows: $\lambda_{id}=0.5$, $\lambda_{per}=10^{-6}$, $\lambda_{edge}=0.2$, $\lambda_g=1$.

\subsection{Predicting Sharp Latent Codes from a Blurry Image}
\label{sec:blurry2MultiZ_training}
In this section we describe how to train a blurry image encoder $E_{v}$ that maps a blurry image $x_i$ to a sequence of 5 latent codes $\bm{z}_i=[z_{i-m}, z_{i-m/2}, z_{i}, z_{i+m/2}, z_{i+m}]$ corresponding to the target sharp frame sequence $\bm{y}_i$.
We train the encoder $E_{v}$ by using the the pre-trained sharp image encoder $E_s$ as teacher. Let $\bm{z}_i=[E_s(y_{i-m}), \ldots, E_s(y_{i+m})]$ denote the sequence of target codes obtained by encoding each target sharp image in the sequence $\bm{y}_i$ with $E_s$. 

Jin~\etal~\cite{Jin_2018_CVPR} point out ambiguities when regressing a sequence of sharp frames from a blurry image. 
Indeed, the order of the regressed frames can be ambiguous since the output sequence is often valid whether it is played forward or backward. 
We handle this forward/backward ambiguity by allowing for either solution in the training objective. 
Let the reversed target sequence be denoted with $\bm{\Bar{z}}_i=[E_s(y_{i+m}), \ldots, E_s(y_{i-m})]$.
The training objective for $E_{v}$ is then given by
\begin{equation}
\min_{E_v}\sum_{n=1}^{n} \min \left( \left| E_{v}(x_{i}) - \bm{z}_i  \right|, \left| E_{v}(x_{i}) - \bm{\Bar{z}}_i  \right| \right), 
\end{equation}
where we minimize either over the forward or backward target sequence, depending on which one better matches the prediction. 

\subsection{Regressing a 3D Face Model}
\label{sec:3dvideo_training}
To perform a novel view synthesis of the reconstructed sharp frame sequence, we need to know the 3D rotation of the face. 
Our approach is to learn to extract the 3D viewpoint of a face by training an encoder $E_{3D}$ to regress the coefficients of a 3DMM \cite{bfm09} along with camera parameters that define the rotation angles $R \in \mathbb{R}^{3}$, the translation $t \in \mathbb{R}^{3}$, and the illumination coefficients $\gamma \in \mathbb{R}^{9}$.
The 3DMM coefficients can be grouped into components responsible for representing identity $\alpha$, texture $\beta$, and facial expression $\delta$. 
Given a blurry face image $x_i^\nu$ from view $\nu$, we thus train a ResNet-50 \cite{He_2016_CVPR} to regress the vector $c_i^\nu=(\alpha_i, \beta_i, \delta_i, \gamma_i^\nu,  R_i^\nu,  t_i^\nu) \in \mathbb{R}^{460}$ of 3D coefficients corresponding to the sharp middle frame $y_i^\nu$. 
The predicted 3D coefficients $c_i^\nu$ are passed through a differentiable renderer $\phi$ \cite{szabo2019unsupervised} and the 3D encoder $E_{3D}$ is trained by minimizing
\begin{equation}
    \min_{E_{3D}}\sum_{i=1}^{n} \sum_{\nu=1}^{\nu_{i}} \ell_{im} \left( \phi \left(E_{3D}(x_i^\nu) \right), y_i^\nu \right) + \ell_{3D} \left( E_{3D}(x_i^\nu), y_i^\nu \right) + \lambda_c ( |\alpha_i|^2+ |\beta_i|^2 + |\delta_i|^2 ),
\end{equation}
where $\ell_{im}$ and $\ell_{3D}$ are a combination of different reconstruction losses (see supplementary for details), and $\lambda_c=10^{-4}$ controls the amount of regularization applied to the 3DMM coefficients to prevent a degradation of face shape and texture.
Note that the coefficients, $\alpha, \beta, \gamma$, are shared across different views, promoting the accurate learning of facial expressions.

\subsection{Learning to Rotate Faces in Latent Space }
\label{sec:fusion_training}
Given a blurry image $x_i^\nu$ from viewpoint $\nu$ and associated latent codes $\bm{z}_i^\nu=E_v(x_i^\nu)$ as well as pose information $E_{3D}(x_i^\nu)$, we aim to manipulate $\bm{z}_i^\nu$ in latent space such that the reconstruction exhibits a desired change of viewpoint. 
We implement this by learning a fusion network $F$ that takes as input a pair $(z_j^\nu, \Delta p)$ consisting of a single frame encoding $z_j^\nu$ and a relative change in pose $\Delta p$. 
The output modified latent codes are then given by applying $F$ to all frames in the sequence independently, \ie, the modified codes are given by $\bm{z}_i^{\nu+ \Delta p}=[F(z_{i-m}^\nu, \Delta p), \ldots, F(z_{i+m}^\nu, \Delta p)]$.

During training, we sample two blurry images $x_i^{u}$ and $x_i^{v}$ from two different viewpoints, but with the same timestamp. 
The change in viewpoint is then computed from $E_{3D}(x_i^{u})-E_{3D}(x_i^{v})$, which corresponds to $\Delta p_i^{uv}=(R_i^{v}-R_i^{u})$, \ie, the difference in the estimated 3D rotation angles between the two views.
We train the fusion model $F$ to regress the latent codes $\bm{z}_i^v$ from the pair $(\bm{z}_i^u, \Delta p_i^{uv})$ by optimizing the following objective
\begin{equation}
\min_{F}\sum_{i=1}^{n} \sum_{u\neq v} \min ( \left| F(\bm{z}_i^u, \Delta p_i^{uv})-\bm{z}_i^v \right|, \left| F(\bm{z}_i^u, \Delta p_i^{uv})-\bm{\Bar{z}}_i^v \right| ), 
\end{equation}
where the \texttt{min} function again takes care of possible frame order ambiguities.
\subsection{Implementation Details}
We employed ResNet-50 \cite{He_2016_CVPR} as a backbone architecture for $E_{s}, E_{v}$ and $E_{3d}$. 
The average-pooled features are fed through fully-connected layers with  $14 \times 512$ (single frame), $5 \times 14 \times 512$ (5 frames) and $460$ neurons for $E_{s}, E_{v}$ and $E_{3d}$ respectively. 
The generator G is pre-trained with all hyper-parameters set to their default values on 8 NVIDIA GTX 1080Ti GPUs (see \citet{Karras2019stylegan2} for details). All other networks were trained on 3 NVIDIA GeForce RTX 3090 GPUs. The Adam optimizer \cite{kingma2014adam} with a fixed learning rate of $10^{-4}$ was used for the training of all the networks. We used batch sizes of 72, 96, 90, 84 samples for  $E_{s}, E_{v}, E_{3d}$ and $F$ respectively. We trained our models $E_{s}, E_{v}, E_{3d}$ and $F$ for 1000K, 100K, 600K, and 500K iterations each. The ratio of samples within one batch stemming from FFHQ, 300VW and BMFD is 2:1:1. 
All the models are trained on an image resolution of $256\times256$. We used random jittering of hue, brightness, saturation, and contrast for data augmentation.

\section{Experiments}
\noindent\textbf{Datasets.} Besides our novel multi-view face dataset we also use 300VW~\cite{300VW}, FFHQ~\cite{STYLEGAN} and VIDTIMIT~\cite{VIDTIMIT} in our experiments. To synthesize motion-blurred images for training, we average \textit{(i)} 65 consecutive frames from videos of 40 identities of our new dataset, and \textit{(ii)} 9 consecutive frames from 65 identities of 300VW. 
To increase the number of identities for training and avoid overfitting, we also incorporate samples from FFHQ.
Since FFHQ consists of still images, we simulate blurs by convolving images with randomly sampled $9\times9$ motion blur kernels. 
Because 300VW and FFHQ lack multiple views, we simulate them via horizontal mirroring of frames. 
We evaluate our method on the remaining identities of our new dataset and the VIDTIMIT dataset.\\ 
\noindent \textbf{Pose-Regression Accuracy of $\bm{E_{v}}$.}
We perform experiments to quantify the facial pose accuracy of the reconstructed frame sequence $G(E_{v}(x))$. 
To this end, we extract facial landmarks using the method of \cite{bulat2017far} from both the reconstructed and the ground-truth frame sequence on test subjects of our dataset.
We report the MSE between them in Table~\ref{tab:blurry2sequence_landmarks} (again adjusting for the forward/backward ambiguity).
We observe that the mean landmark error is slightly larger for peripheral frames (1, 2, 4, and 5) than the middle one (3). 
The mean landmark error is 3.46 pixels which amounts to 1.35\% of the $256\times256$ image resolution.\\
\noindent \textbf{Identity Preservation and Pose Accuracy under Novel View Synthesis.} 
A key component of our method is the fusion model $F$, which performs the manipulation in the latent space that results in a change of the viewpoint.
We thus perform ablation experiments for different architecture designs of $F$, where we measure how well they reconstruct the pose in novel views and how well they preserve the identity of the face. 
We consider two functional designs: \textit{(i)} $FCxR$, where $F$ is modelled via residual computation, \ie, $F(z, \Delta p)=z+MLP_x([z, \Delta p])$, and  \textit{(ii)} $FCx$, where $F$ simply consists of $x$ fully-connected layers, \ie, $F(z, \Delta p)=MLP_x([z, \Delta p])$ ($x$ indicates the number of layers in the $MLP$).
We want $F$ only to affect the 3D orientation of the face in our method and preserve the face identity as much as possible. 
To quantify the consistency of face identities under novel view synthesis, we compute the agreement of a pre-trained identity classifier \cite{VGGFACE2} between a restored frontal view and reconstructions under varying amounts of rotation.  
We report the resulting Top-1 and Top-5 label agreements on VIDTIMIT in Table~\ref{tab:rot_identity_agreement}.
Because the identity classifier is not perfectly robust to face rotations, we also report the estimated identity agreement of the classifier (its sensitivity) on sharp ground-truth rotations. 
We observe that the identity labels of rotated sequences are relatively consistent with the classifier's sensitivity on ground truth rotations up to $\pm30^{\circ}$. The residual version $FCxR$ performs considerably better. 
\begin{table}[!t]
\begin{minipage}{1.0\textwidth}
\begin{minipage}[!t]{0.47\textwidth}
\centering
\small
\setlength{\tabcolsep}{2.2pt}
\renewcommand{\arraystretch}{0.85}
\begin{tabular}{l@{\hspace{1.5mm}}c@{\hspace{1.5mm}}c@{\hspace{1.5mm}}c@{\hspace{1.5mm}}c@{\hspace{1.5mm}}c}\toprule
\multicolumn{1}{l}{\multirow{2}{*}{\textbf{Frames}}} & \multicolumn{5}{c}{\textbf{Views}}\\
\multicolumn{1}{l}{} & \multicolumn{1}{c}{1,8} & \multicolumn{1}{c}{2,7} & \multicolumn{1}{c}{3,6} & \multicolumn{1}{c}{4,5} & \multicolumn{1}{c}{All}\\\midrule

\multicolumn{1}{l}{Middle 3} & \multicolumn{1}{c}{2.93} & \multicolumn{1}{c}{2.78} & \multicolumn{1}{c}{2.89} & \multicolumn{1}{c}{2.82} & \multicolumn{1}{c}{2.85}\\

\multicolumn{1}{l}{Frames 2,4} & \multicolumn{1}{c}{3.31} & \multicolumn{1}{c}{3.13} & \multicolumn{1}{c}{3.29} & \multicolumn{1}{c}{3.27} & \multicolumn{1}{c}{3.25}\\

\multicolumn{1}{l}{Frames 1,5} & \multicolumn{1}{c}{3.94} & \multicolumn{1}{c}{3.84} & \multicolumn{1}{c}{3.99} & \multicolumn{1}{c}{4.01} & \multicolumn{1}{c}{3.95}\\

\multicolumn{1}{l}{All Frames} & \multicolumn{1}{c}{3.50} & \multicolumn{1}{c}{3.35} & \multicolumn{1}{c}{3.51} & \multicolumn{1}{c}{3.49} & \multicolumn{1}{c}{3.46}\\ \bottomrule
\end{tabular}
\caption{
\textbf{Same view landmark error.}
We report the landmark error (in pixels) between the ground-truth and reconstructed frame sequences without rotation.
\label{tab:blurry2sequence_landmarks}}
\end{minipage}
\hfill
\begin{minipage}[!t]{0.5\textwidth}
\centering
\small
\setlength{\tabcolsep}{1.0pt}
\renewcommand{\arraystretch}{0.8}
\begin{tabular}{l@{\hspace{4mm}}c@{\hspace{4mm}}c@{\hspace{3.5mm}}c}\toprule
\multirow{2}{*}{\textbf{Fusion}} & \multicolumn{3}{c}{\textbf{Viewpoint Change}} \\
  & $\pm30^{\circ}$ & $\pm45^{\circ}$ & $\pm60^{\circ}$ \\ \midrule
FC3   & 51\%~(86\%) & 27\%~(62\%) & 14\%~(37\%) \\
FC3R    & 56\%~(84\%) & 36\%~(66\%) & 22\%~(45\%) \\\bottomrule
\end{tabular}
\caption{
\textbf{Identity agreement between frontal and rotated sequences.} We report the Top-1~(Top-5) label agreement of a pre-trained identity classifier between frontal and rotated views. Note that the classifier has a sensitivity of 61\%~(87\%) on average over all viewpoints. 
\label{tab:rot_identity_agreement}}
\end{minipage}
\end{minipage}
\vspace{-10pt}
\end{table}
\begin{figure*}[t]
\centering
\includegraphics[width=0.95\textwidth]{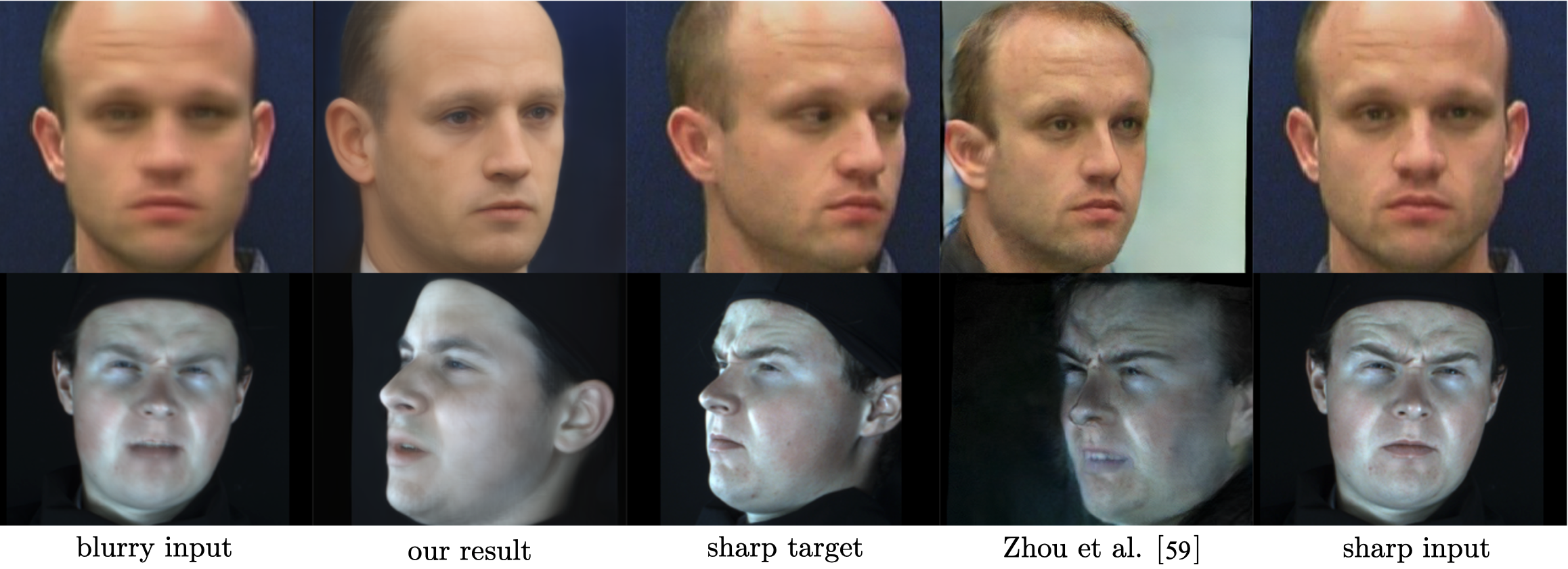}
\caption{
\textbf{Qualitative novel view comparison to \citet{Zhou_2020_CVPR}.}
We compare on VIDTIMIT (top) and BMFD (bottom). Note that \cite{Zhou_2020_CVPR} predicts novel views from the sharp input image on the right, whereas we predict it from the blurry image on the left. 
\label{fig:novel_view_compatitor_qualitative}}
\vspace{-10pt}
\end{figure*}

\begin{table}[!t]
\begin{minipage}{1.0\textwidth}
\begin{minipage}[!t]{0.5\textwidth}
\centering
\small
\setlength{\tabcolsep}{3.0pt}
\renewcommand{\arraystretch}{0.8}
\begin{tabular}{l@{\hspace{1.0mm}}l@{\hspace{1.0mm}}c@{\hspace{1.0mm}}c@{\hspace{1.0mm}}c@{\hspace{1.0mm}}c@{\hspace{1.0mm}}c}\toprule
\multicolumn{1}{l}{\multirow{2}{*}{\textbf{Frames}}} & \multicolumn{1}{l}{\multirow{2}{*}{\textbf{Fusion}}} & \multicolumn{5}{c}{\textbf{Views}}\\
\multicolumn{1}{l}{} & \multicolumn{1}{l}{} & \multicolumn{1}{c}{1,8} & \multicolumn{1}{c}{2,7} & \multicolumn{1}{c}{3,6} & \multicolumn{1}{c}{4,5} & \multicolumn{1}{c}{All}\\\midrule

\multicolumn{1}{l}{\multirow{2}{*}{Middle 3}} & \multicolumn{1}{l}{FC3R} & \multicolumn{1}{c}{6.07} & \multicolumn{1}{c}{7.37} & \multicolumn{1}{l}{7.03} & \multicolumn{1}{c}{3.80} & \multicolumn{1}{c}{6.07} \\
           
\multicolumn{1}{l}{} & \multicolumn{1}{l}{FC3} & \multicolumn{1}{c}{6.67} & \multicolumn{1}{c}{7.51} & \multicolumn{1}{c}{7.02} & \multicolumn{1}{l}{3.99} & \multicolumn{1}{c}{6.30} \\\midrule
            
\multicolumn{1}{l}{\multirow{2}{*}{Frames 2,4 }} & \multicolumn{1}{l}{FC3R} & \multicolumn{1}{c}{6.08} & \multicolumn{1}{c}{7.33} & \multicolumn{1}{l}{7.03} & \multicolumn{1}{c}{3.85} & \multicolumn{1}{c}{6.07} \\
            
\multicolumn{1}{l}{} & \multicolumn{1}{l}{FC3 } & \multicolumn{1}{c}{6.61} & \multicolumn{1}{c}{7.49} & \multicolumn{1}{c}{6.99} & \multicolumn{1}{l}{4.02} & \multicolumn{1}{c}{6.28} \\\midrule

\multicolumn{1}{l}{\multirow{2}{*}{Frames 1,5 }} & \multicolumn{1}{l}{FC3R} & \multicolumn{1}{c}{6.20} & \multicolumn{1}{c}{7.46} & \multicolumn{1}{l}{7.17} & \multicolumn{1}{c}{4.03} & \multicolumn{1}{c}{6.21} \\
            
\multicolumn{1}{l}{} & \multicolumn{1}{l}{FC3} & \multicolumn{1}{c}{6.63} & \multicolumn{1}{c}{7.63} & \multicolumn{1}{c}{7.14} & \multicolumn{1}{l}{4.20} & \multicolumn{1}{c}{6.40} \\\midrule
            
\multicolumn{1}{l}{\multirow{2}{*}{All Frames}} & \multicolumn{1}{l}{FC3R} & \multicolumn{1}{c}{6.12} & \multicolumn{1}{c}{7.39} & \multicolumn{1}{l}{7.09} & \multicolumn{1}{c}{3.91} & \multicolumn{1}{c}{6.13} \\
            
\multicolumn{1}{l}{} & \multicolumn{1}{l}{FC3} & \multicolumn{1}{c}{6.63} & \multicolumn{1}{c}{7.55} & \multicolumn{1}{c}{7.06} & \multicolumn{1}{l}{4.09} & \multicolumn{1}{c}{6.33} \\\bottomrule
\end{tabular}
\end{minipage}
\begin{minipage}{0.49\textwidth}
\centering
\includegraphics[width=1.0\textwidth]{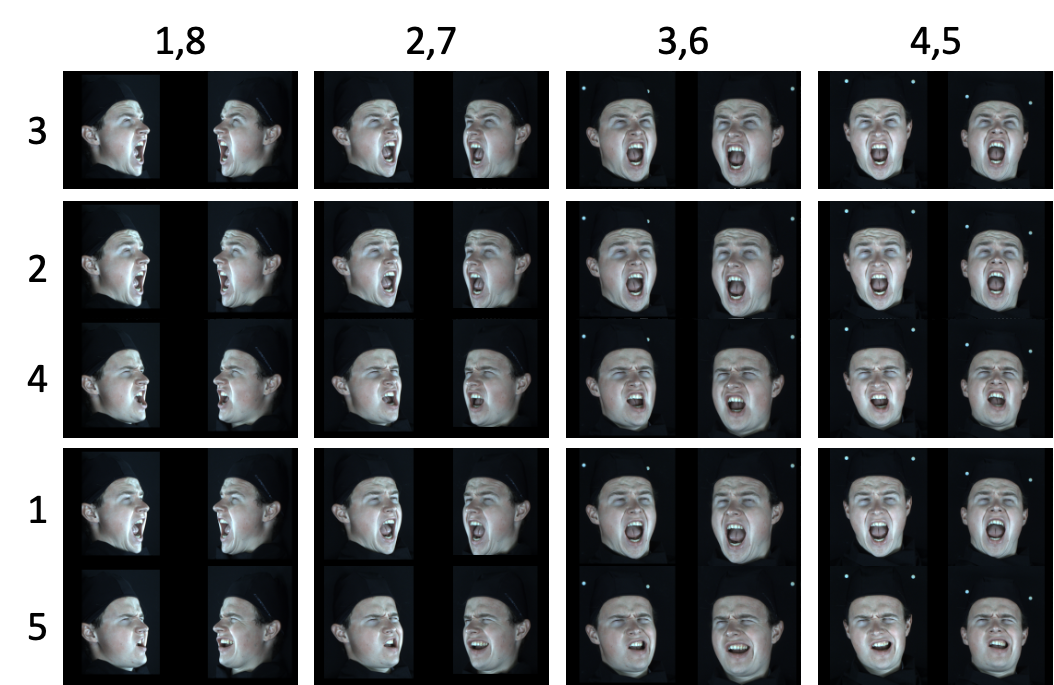}
\end{minipage}
\caption{
\textbf{Face landmark accuracy for different fusion models.}
In the table we report the landmark error of different frames in the reconstructed sequence (rows) and when faces are rotated to the different views in BMFD (columns). The blurry input image is taken from view 4 in all cases. 
An illustration of the frame and view layout is given on the right. 
\label{tab:fusion_landmarks}}
\end{minipage}
\vspace{-20pt}
\end{table}

\begin{figure*}[t]
\vspace{-3pt}
\centering
\includegraphics[width=1.0\textwidth]{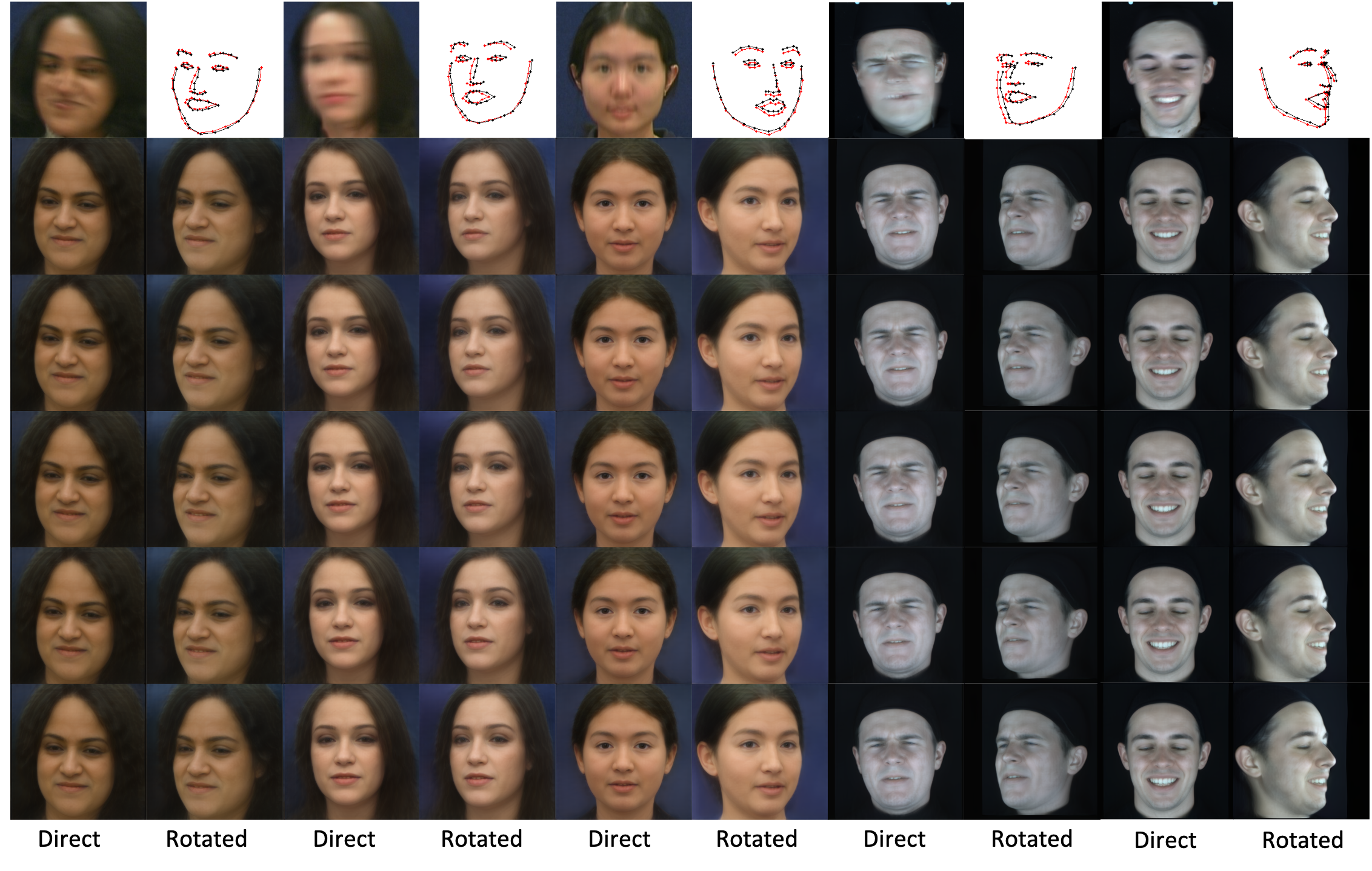}
\vspace{-15pt}
\caption{
\textbf{Sample sharp video reconstructions from our model.} We show reconstructed frame sequences without viewpoint change (odd columns) and with random viewpoint changes (even columns). 
The first row shows the blurry input image followed by landmarks computed on the first and last frame in the reconstructed sequence.
The first three examples are computed on VIDTIMIT and the last two on our test set. 
\label{fig:qualitative}}
\vspace{-13pt}
\end{figure*}

To quantify the accuracy of the predicted face pose under novel view synthesis, we measure the face landmark error between the ground truth views and our reconstructions on test subjects of our multi-view dataset. 
Blurry frontal images (view 4) are fed through our model to reconstruct sharp frame sequences corresponding to the other seven views in our dataset.
We report the mean landmark errors of different fusion models for all the views and predicted frames in Table~\ref{tab:fusion_landmarks}. 
We observe that the average error across all views and frames varies between 6.13 and 6.33 pixels. 
Note that the reconstructions without rotations already show a mean landmark error of 3.46 pixels (see Table~\ref{tab:blurry2sequence_landmarks}).
Qualitative reconstructions of frontal and rotated frame sequences obtained with our method can be found in Figure~\ref{fig:qualitative}.

\noindent \textbf{Comparison to Prior Work.}
We compare to \citet{Zhou_2020_CVPR} on novel face view synthesis quantitatively in Table~\ref{tab:comparison_landmarks} and qualitatively in Figure~\ref{fig:novel_view_compatitor_qualitative}.
Since \cite{Zhou_2020_CVPR} is trained on non-blurry face images, we feed it with sharp frontal views from VIDTIMIT and our test set. 
Our method was instead evaluated on blurry input images. Despite this disadvantage, our method yields a comparable accuracy. More results are shown in the supplemental material.\\
We evaluated the performance of our system using conventional metrics such as PSNR and SSIM. None of the existing prior deblurring work can generate novel views from a blurry input. Therefore, we use the combination of two methods for comparison purposes. We extract the sharp video sequence from a blurry input utilizing the method of \citet{Jin_2018_CVPR} and subsequently rotate the resulting frames using the method of \citet{Zhou_2020_CVPR}. The mean PSNR and SSIM between ground-truth and rotated sequences are reported in Table~\ref{tab:comparison_psnr_ssim}.

\begin{table}[!t]
\centering
\small
\setlength{\tabcolsep}{5.0pt}
\renewcommand{\arraystretch}{0.8}
\begin{tabular}{l@{\hspace{1.5mm}}c@{\hspace{1.5mm}}c@{\hspace{1.5mm}}c@{\hspace{1.5mm}}c@{\hspace{1.5mm}}c@{\hspace{1.5mm}}c}\toprule
\multicolumn{1}{l}{\multirow{2}{*}{\textbf{Method}}} & \multicolumn{5}{c}{\textbf{BMFD}} & \multicolumn{1}{c}{\multirow{2}{*}{\textbf{VIDTIMIT}}}\\
\multicolumn{1}{l}{} & \multicolumn{1}{c}{1,8} & \multicolumn{1}{c}{2,7} & \multicolumn{1}{c}{3,6} & \multicolumn{1}{c}{4,5} & \multicolumn{1}{c}{All} & \multicolumn{1}{c}{}\\\midrule

\multicolumn{1}{l}{\citet{Zhou_2020_CVPR}} & \multicolumn{1}{c}{7.12} & \multicolumn{1}{c}{6.42} & \multicolumn{1}{c}{5.40} & \multicolumn{1}{c}{5.61} & \multicolumn{1}{c}{6.14} & \multicolumn{1}{c}{3.12}\\
\multicolumn{1}{l}{Ours} & \multicolumn{1}{c}{6.07} & \multicolumn{1}{c}{7.37} & \multicolumn{1}{c}{7.03} & \multicolumn{1}{c}{3.80} & \multicolumn{1}{c}{6.07} & \multicolumn{1}{c}{3.96}\\ \bottomrule
\end{tabular}
\caption{
\textbf{Novel view pose error comparison.}
We compare to the prior novel face view synthesis method by \cite{Zhou_2020_CVPR} in terms of face landmark accuracy on VIDTIMIT and BMFD.
\label{tab:comparison_landmarks}}
\end{table}

\begin{table}[!t]
\vspace{-12pt}
\centering
\small
\setlength{\tabcolsep}{5.0pt}
\renewcommand{\arraystretch}{0.8}
\begin{tabular}{l@{\hspace{1.5mm}}c@{\hspace{1.5mm}}c@{\hspace{1.5mm}}}\toprule
\multicolumn{1}{l}{\textbf{Method}} & \multicolumn{1}{c}{\textbf{PSNR}} & \multicolumn{1}{c}{\textbf{SSIM}}\\\midrule

\multicolumn{1}{l}{\citet{Jin_2018_CVPR} + \citet{Zhou_2020_CVPR}} & \multicolumn{1}{c}{16.07} & \multicolumn{1}{c}{0.38} \\
\multicolumn{1}{l}{Ours} & \multicolumn{1}{c}{19.45} & \multicolumn{1}{c}{0.60} \\ \bottomrule
\end{tabular}
\caption{
\textbf{Novel view PSNR and SSIM comparison.}
We compare to the prior work in terms of PSNR and SSIM metrics on our dataset. First, the blurry input images from view 4 are fed to the method of \citet{Jin_2018_CVPR}, then, the resulting deblurred sequences are rotated using the method of \citet{Zhou_2020_CVPR}.
\label{tab:comparison_psnr_ssim}}
\vspace{-15pt}
\end{table}
\section{Conclusions}
In this paper, we have presented a first method to reconstruct novel view videos from a single motion-blurred face image. Capabilities of the method were demonstrated on the VIDTIMIT dataset and a novel high frame rate, multi-view facial dataset, which we introduced. The multi-view dataset is crucial in enabling the training of our model. Moreover, our dataset is not limited to our proposed task: It can also be used to evaluate facial restoration methods for 3D reconstruction, single/video super-resolution, and temporal frame interpolation.\\
\noindent \textbf{Acknowledgements.} This work was supported by grant $200021$\_$165845$ of the Swiss National Science Foundation.
\newpage
\bibliography{main}

\begin{thebibliography}{60}
\providecommand{\natexlab}[1]{#1}
\providecommand{\url}[1]{\texttt{#1}}
\expandafter\ifx\csname urlstyle\endcsname\relax
  \providecommand{\doi}[1]{doi: #1}\else
  \providecommand{\doi}{doi: \begingroup \urlstyle{rm}\Url}\fi

\bibitem[ski(2002)]{skin_mask_ref}
Statistical color models with application to skin detection.
\newblock \emph{International Journal of Computer Vision}, 46\penalty0 (1),
  2002.

\bibitem[Blanz and Vetter(1999)]{Blanz_1999_3DMM}
Volker Blanz and Thomas Vetter.
\newblock A morphable model for the synthesis of 3d faces.
\newblock In \emph{SIGGRAPH}, 1999.

\bibitem[Bolkart and Wuhrer(2016)]{Bolkart_2016_CVPR}
Timo Bolkart and Stefanie Wuhrer.
\newblock A robust multilinear model learning framework for 3d faces.
\newblock In \emph{CVPR}, 2016.

\bibitem[Booth et~al.(2016)Booth, Roussos, Zafeiriou, Ponniah, and
  Dunaway]{Booth_2016_CVPR}
James Booth, Anastasios Roussos, Stefanos Zafeiriou, Allan Ponniah, and David
  Dunaway.
\newblock A 3d morphable model learnt from 10,000 faces.
\newblock In \emph{CVPR}, 2016.

\bibitem[Booth et~al.(2017)Booth, Antonakos, Ploumpis, Trigeorgis, Panagakis,
  and Zafeiriou]{Booth_2017_CVPR}
James Booth, Epameinondas Antonakos, Stylianos Ploumpis, George Trigeorgis,
  Yannis Panagakis, and Stefanos Zafeiriou.
\newblock 3d face morphable models "in-the-wild".
\newblock In \emph{CVPR}, 2017.

\bibitem[Bulat and Tzimiropoulos(2017)]{bulat2017far}
Adrian Bulat and Georgios Tzimiropoulos.
\newblock How far are we from solving the 2d \& 3d face alignment problem? (and
  a dataset of 230,000 3d facial landmarks).
\newblock In \emph{International Conference on Computer Vision}, 2017.

\bibitem[Cao et~al.(2018)Cao, Shen, Xie, Parkhi, and Zisserman]{VGGFACE2}
Qiong Cao, Li~Shen, Weidi Xie, Omkar~M. Parkhi, and Andrew Zisserman.
\newblock {VGGFace2}: A dataset for recognising faces across pose and age.
\newblock In \emph{International Conference on Automatic Face and Gesture
  Recognition}, 2018.

\bibitem[Chrysos and Zafeiriou(2017)]{Chrysos_2017_CVPR_Workshops}
Grigorios~G. Chrysos and Stefanos Zafeiriou.
\newblock Deep face deblurring.
\newblock In \emph{Proceedings of the IEEE Conference on Computer Vision and
  Pattern Recognition (CVPR) Workshops}, July 2017.

\bibitem[Chrysos et~al.(2019)Chrysos, Favaro, and
  Zafeiriou]{chrysos_face_deblur}
Grigorios~G. Chrysos, Paolo Favaro, and Stefanos Zafeiriou.
\newblock Motion deblurring of faces.
\newblock \emph{International Journal of Computer Vision}, 127\penalty0 (6),
  2019.

\bibitem[Chrysos et~al.(2015)Chrysos, Antonakos, Zafeiriou, and Snape]{300VW}
Grigoris~G. Chrysos, Epameinondas Antonakos, Stefanos Zafeiriou, and Patrick
  Snape.
\newblock Offline deformable face tracking in arbitrary videos.
\newblock In \emph{2015 IEEE International Conference on Computer Vision
  Workshop (ICCVW)}, 2015.

\bibitem[Egger et~al.(2020)Egger, Smith, Tewari, Wuhrer, Zollhoefer, Beeler,
  Bernard, Bolkart, Kortylewski, Romdhani, Theobalt, Blanz, and
  Vetter]{egger20203d}
Bernhard Egger, William A.~P. Smith, Ayush Tewari, Stefanie Wuhrer, Michael
  Zollhoefer, Thabo Beeler, Florian Bernard, Timo Bolkart, Adam Kortylewski,
  Sami Romdhani, Christian Theobalt, Volker Blanz, and Thomas Vetter.
\newblock 3d morphable face models - past, present and future.
\newblock \emph{ACM Transactions on Graphics}, 39\penalty0 (5), August 2020.

\bibitem[Genova et~al.(2018)Genova, Cole, Maschinot, Sarna, Vlasic, and
  Freeman]{Genova_2018_CVPR}
Kyle Genova, Forrester Cole, Aaron Maschinot, Aaron Sarna, Daniel Vlasic, and
  William~T. Freeman.
\newblock Unsupervised training for 3d morphable model regression.
\newblock In \emph{CVPR}, 2018.

\bibitem[Hassner et~al.(2015)Hassner, Harel, Paz, and
  Enbar]{hassner2015effective}
Tal Hassner, Shai Harel, Eran Paz, and Roee Enbar.
\newblock Effective face frontalization in unconstrained images.
\newblock In \emph{Proceedings of the IEEE conference on computer vision and
  pattern recognition}, 2015.

\bibitem[He et~al.(2016)He, Zhang, Ren, and Sun]{He_2016_CVPR}
Kaiming He, Xiangyu Zhang, Shaoqing Ren, and Jian Sun.
\newblock Deep residual learning for image recognition.
\newblock In \emph{CVPR}, 2016.

\bibitem[Hu et~al.(2018)Hu, Wu, Yu, He, and Sun]{hu2018pose}
Yibo Hu, Xiang Wu, Bing Yu, Ran He, and Zhenan Sun.
\newblock Pose-guided photorealistic face rotation.
\newblock In \emph{Proceedings of the IEEE conference on computer vision and
  pattern recognition}, 2018.

\bibitem[Huang et~al.(2017)Huang, Zhang, Li, and He]{huang2017beyond}
Rui Huang, Shu Zhang, Tianyu Li, and Ran He.
\newblock Beyond face rotation: Global and local perception gan for
  photorealistic and identity preserving frontal view synthesis.
\newblock In \emph{Proceedings of the IEEE International Conference on Computer
  Vision}, 2017.

\bibitem[Jin et~al.(2018{\natexlab{a}})Jin, Hirsch, and
  Favaro]{Jin_2018_CVPR_Workshops}
Meiguang Jin, Michael Hirsch, and Paolo Favaro.
\newblock Learning face deblurring fast and wide.
\newblock In \emph{Proceedings of the IEEE Conference on Computer Vision and
  Pattern Recognition (CVPR) Workshops}, June 2018{\natexlab{a}}.

\bibitem[Jin et~al.(2018{\natexlab{b}})Jin, Meishvili, and
  Favaro]{Jin_2018_CVPR}
Meiguang Jin, Givi Meishvili, and Paolo Favaro.
\newblock Learning to extract a video sequence from a single motion-blurred
  image.
\newblock In \emph{CVPR}, 2018{\natexlab{b}}.

\bibitem[Jin et~al.(2019)Jin, Hu, and Favaro]{Jin_2019_CVPR}
Meiguang Jin, Zhe Hu, and Paolo Favaro.
\newblock Learning to extract flawless slow motion from blurry videos.
\newblock In \emph{CVPR}, 2019.

\bibitem[Karras et~al.(2019)Karras, Laine, and Aila]{STYLEGAN}
Tero Karras, Samuli Laine, and Timo Aila.
\newblock A style-based generator architecture for generative adversarial
  networks.
\newblock In \emph{2019 IEEE/CVF Conference on Computer Vision and Pattern
  Recognition (CVPR)}, 2019.

\bibitem[Karras et~al.(2020)Karras, Laine, Aittala, Hellsten, Lehtinen, and
  Aila]{Karras2019stylegan2}
Tero Karras, Samuli Laine, Miika Aittala, Janne Hellsten, Jaakko Lehtinen, and
  Timo Aila.
\newblock Analyzing and improving the image quality of {StyleGAN}.
\newblock In \emph{Proc. CVPR}, 2020.

\bibitem[Kato et~al.(2018)Kato, Ushiku, and Harada]{kato2018neural}
Hiroharu Kato, Yoshitaka Ushiku, and Tatsuya Harada.
\newblock Neural 3d mesh renderer.
\newblock In \emph{CVPR}, 2018.

\bibitem[Kim et~al.(2018)Kim, Zollhöfer, Tewari, Thies, Richardt, and
  Theobalt]{Kim_2018_CVPR}
Hyeongwoo Kim, Michael Zollhöfer, Ayush Tewari, Justus Thies, Christian
  Richardt, and Christian Theobalt.
\newblock Inversefacenet: Deep monocular inverse face rendering.
\newblock In \emph{CVPR}, 2018.

\bibitem[Kingma and Ba(2014)]{kingma2014adam}
Diederik~P Kingma and Jimmy Ba.
\newblock Adam: A method for stochastic optimization.
\newblock \emph{arXiv preprint arXiv:1412.6980}, 2014.

\bibitem[Liu et~al.(2019)Liu, Tran, and Liu]{Liu_2019_ICCV}
Feng Liu, Luan Tran, and Xiaoming Liu.
\newblock 3d face modeling from diverse raw scan data.
\newblock In \emph{ICCV}, 2019.

\bibitem[Lu et~al.(2019)Lu, Chen, and Chellappa]{Lu_2019_CVPR}
Boyu Lu, Jun-Cheng Chen, and Rama Chellappa.
\newblock Unsupervised domain-specific deblurring via disentangled
  representations.
\newblock In \emph{CVPR}, 2019.

\bibitem[Masi et~al.(2018)Masi, Wu, Hassner, and Natarajan]{masi2018deep}
Iacopo Masi, Yue Wu, Tal Hassner, and Prem Natarajan.
\newblock Deep face recognition: A survey.
\newblock In \emph{2018 31st SIBGRAPI conference on graphics, patterns and
  images (SIBGRAPI)}. IEEE, 2018.

\bibitem[Meishvili et~al.(2020)Meishvili, Jenni, and
  Favaro]{Meishvili_2020_CVPR}
Givi Meishvili, Simon Jenni, and Paolo Favaro.
\newblock Learning to have an ear for face super-resolution.
\newblock In \emph{IEEE/CVF Conference on Computer Vision and Pattern
  Recognition (CVPR)}, June 2020.

\bibitem[Nah et~al.(2017)Nah, Hyun~Kim, and Mu~Lee]{Nah_2017_CVPR}
Seungjun Nah, Tae Hyun~Kim, and Kyoung Mu~Lee.
\newblock Deep multi-scale convolutional neural network for dynamic scene
  deblurring.
\newblock In \emph{CVPR}, 2017.

\bibitem[Pan et~al.(2014)Pan, Hu, Su, and Yang]{Pan_ECCV_2014_face_deblur}
Jinshan Pan, Zhe Hu, Zhixun Su, and Ming-Hsuan Yang.
\newblock Deblurring face images with exemplars.
\newblock In \emph{ECCV}, 2014.

\bibitem[Paysan et~al.(2009)Paysan, Knothe, Amberg, Romdhani, and
  Vetter]{bfm09}
P.~Paysan, R.~Knothe, B.~Amberg, S.~Romdhani, and T.~Vetter.
\newblock A 3d face model for pose and illumination invariant face recognition.
\newblock In \emph{AVSS}, 2009.

\bibitem[Peng et~al.(2017)Peng, Feng, Xu, and Su]{Peng_2017_CVPR}
Weilong Peng, Zhiyong Feng, Chao Xu, and Yong Su.
\newblock Parametric t-spline face morphable model for detailed fitting in
  shape subspace.
\newblock In \emph{CVPR}, 2017.

\bibitem[Piotraschke and Blanz(2016)]{Piotraschke_2016_CVPR}
Marcel Piotraschke and Volker Blanz.
\newblock Automated 3d face reconstruction from multiple images using quality
  measures.
\newblock In \emph{CVPR}, 2016.

\bibitem[Ploumpis et~al.(2019)Ploumpis, Wang, Pears, Smith, and
  Zafeiriou]{Ploumpis_2019_CVPR}
Stylianos Ploumpis, Haoyang Wang, Nick Pears, William A.~P. Smith, and Stefanos
  Zafeiriou.
\newblock Combining 3d morphable models: A large scale face-and-head model.
\newblock In \emph{CVPR}, 2019.

\bibitem[Purohit et~al.(2019)Purohit, Shah, and Rajagopalan]{Purohit_2019_CVPR}
Kuldeep Purohit, Anshul Shah, and A.~N. Rajagopalan.
\newblock Bringing alive blurred moments.
\newblock In \emph{CVPR}, 2019.

\bibitem[Ranjan et~al.(2018)Ranjan, Bolkart, Sanyal, and
  Black]{Ranjan_2018_ECCV}
Anurag Ranjan, Timo Bolkart, Soubhik Sanyal, and Michael~J. Black.
\newblock Generating 3d faces using convolutional mesh autoencoders.
\newblock In \emph{ECCV}, 2018.

\bibitem[Ren et~al.(2019)Ren, Yang, Deng, Wipf, Cao, and Tong]{Ren_2019_ICCV}
Wenqi Ren, Jiaolong Yang, Senyou Deng, David Wipf, Xiaochun Cao, and Xin Tong.
\newblock Face video deblurring using 3d facial priors.
\newblock In \emph{ICCV}, 2019.

\bibitem[Roth et~al.(2016)Roth, Tong, and Liu]{Roth_2016_CVPR}
Joseph Roth, Yiying Tong, and Xiaoming Liu.
\newblock Adaptive 3d face reconstruction from unconstrained photo collections.
\newblock In \emph{CVPR}, 2016.

\bibitem[Sanderson and Lovell(2009)]{VIDTIMIT}
Conrad Sanderson and Brian~C. Lovell.
\newblock Multi-region probabilistic histograms for robust and scalable
  identity inference.
\newblock In Massimo Tistarelli and Mark~S. Nixon, editors, \emph{Advances in
  Biometrics}, Berlin, Heidelberg, 2009. Springer Berlin Heidelberg.

\bibitem[Sanyal et~al.(2019)Sanyal, Bolkart, Feng, and Black]{Sanyal_2019_CVPR}
Soubhik Sanyal, Timo Bolkart, Haiwen Feng, and Michael~J. Black.
\newblock Learning to regress 3d face shape and expression from an image
  without 3d supervision.
\newblock In \emph{CVPR}, 2019.

\bibitem[Shen et~al.(2015)Shen, Zafeiriou, Chrysos, Kossaifi, Tzimiropoulos,
  and Pantic]{Shen_2015_ICCV_Workshops}
Jie Shen, Stefanos Zafeiriou, Grigoris~G. Chrysos, Jean Kossaifi, Georgios
  Tzimiropoulos, and Maja Pantic.
\newblock The first facial landmark tracking in-the-wild challenge: Benchmark
  and results.
\newblock In \emph{Proceedings of the IEEE International Conference on Computer
  Vision (ICCV) Workshops}, December 2015.

\bibitem[Shen et~al.(2018)Shen, Lai, Xu, Kautz, and Yang]{Shen_2018_CVPR}
Ziyi Shen, Wei-Sheng Lai, Tingfa Xu, Jan Kautz, and Ming-Hsuan Yang.
\newblock Deep semantic face deblurring.
\newblock In \emph{CVPR}, 2018.

\bibitem[Simonyan and Zisserman(2014)]{vgg2014}
Karen Simonyan and Andrew Zisserman.
\newblock Very deep convolutional networks for large-scale image recognition.
\newblock \emph{arXiv preprint arXiv:1409.1556}, 2014.

\bibitem[Song et~al.(2019)Song, Zhang, Gong, He, Bao, Pan, Yang, and
  Yang]{SONG_IJCV_2019_FHD}
Yibing Song, Jiawei Zhang, Lijun Gong, Shengfeng He, Linchao Bao, Jinshan Pan,
  Qingxiong Yang, and Ming-Hsuan Yang.
\newblock Joint face hallucination and deblurring via structure generation and
  detail enhancement.
\newblock \emph{International Journal of Computer Vision}, 2019.

\bibitem[Szab{\'o} et~al.(2019)Szab{\'o}, Meishvili, and
  Favaro]{szabo2019unsupervised}
Attila Szab{\'o}, Givi Meishvili, and Paolo Favaro.
\newblock Unsupervised generative 3d shape learning from natural images.
\newblock \emph{arXiv:1910.00287}, 2019.

\bibitem[Tewari et~al.(2018)Tewari, Zollhöfer, Garrido, Bernard, Kim, Pérez,
  and Theobalt]{Tewari_2018_CVPR}
Ayush Tewari, Michael Zollhöfer, Pablo Garrido, Florian Bernard, Hyeongwoo
  Kim, Patrick Pérez, and Christian Theobalt.
\newblock Self-supervised multi-level face model learning for monocular
  reconstruction at over 250 hz.
\newblock In \emph{CVPR}, 2018.

\bibitem[Tewari et~al.(2019)Tewari, Bernard, Garrido, Bharaj, Elgharib, Seidel,
  Perez, Zollhofer, and Theobalt]{Tewari_2019_CVPR}
Ayush Tewari, Florian Bernard, Pablo Garrido, Gaurav Bharaj, Mohamed Elgharib,
  Hans-Peter Seidel, Patrick Perez, Michael Zollhofer, and Christian Theobalt.
\newblock Fml: Face model learning from videos.
\newblock In \emph{CVPR}, 2019.

\bibitem[Tomasello et~al.(2007)Tomasello, Hare, Lehmann, and
  Call]{gaze_importance}
Michael Tomasello, Brian Hare, Hagen Lehmann, and Josep Call.
\newblock Reliance on head versus eyes in the gaze following of great apes and
  human infants: the cooperative eye hypothesis.
\newblock \emph{J Hum Evol}, 52\penalty0 (3), Mar 2007.
\newblock ISSN 0047-2484 (Print); 0047-2484 (Linking).

\bibitem[Tran and Liu(2018)]{Tran_2018_CVPR_1}
Luan Tran and Xiaoming Liu.
\newblock Nonlinear 3d face morphable model.
\newblock In \emph{CVPR}, 2018.

\bibitem[Tran et~al.(2019)Tran, Liu, and Liu]{Tran_2019_CVPR}
Luan Tran, Feng Liu, and Xiaoming Liu.
\newblock Towards high-fidelity nonlinear 3d face morphable model.
\newblock In \emph{CVPR}, 2019.

\bibitem[Voynov and Babenko(2020)]{voynov2020unsupervised}
Andrey Voynov and Artem Babenko.
\newblock Unsupervised discovery of interpretable directions in the gan latent
  space, 2020.

\bibitem[Wu et~al.(2019)Wu, Bao, Chen, Ling, Song, Li, Ngan, and
  Liu]{Wu_2019_CVPR}
Fanzi Wu, Linchao Bao, Yajing Chen, Yonggen Ling, Yibing Song, Songnan Li,
  King~Ngi Ngan, and Wei Liu.
\newblock Mvf-net: Multi-view 3d face morphable model regression.
\newblock In \emph{CVPR}, 2019.

\bibitem[Xu et~al.(2017)Xu, Sun, Pan, Zhang, Pfister, and Yang]{Xu_2017_ICCV_1}
Xiangyu Xu, Deqing Sun, Jinshan Pan, Yujin Zhang, Hanspeter Pfister, and
  Ming-Hsuan Yang.
\newblock Learning to super-resolve blurry face and text images.
\newblock In \emph{Proceedings of the IEEE International Conference on Computer
  Vision (ICCV)}, Oct 2017.

\bibitem[Xu et~al.(2019)Xu, Chen, and Jia]{xu2019view}
Xiaogang Xu, Ying-Cong Chen, and Jiaya Jia.
\newblock View independent generative adversarial network for novel view
  synthesis.
\newblock In \emph{Proceedings of the IEEE International Conference on Computer
  Vision}, 2019.

\bibitem[Yang et~al.(2020)Yang, Zhu, Wang, Huang, Shen, Yang, and
  Cao]{Yang_2020_CVPR}
Haotian Yang, Hao Zhu, Yanru Wang, Mingkai Huang, Qiu Shen, Ruigang Yang, and
  Xun Cao.
\newblock Facescape: A large-scale high quality 3d face dataset and detailed
  riggable 3d face prediction.
\newblock In \emph{CVPR}, 2020.

\bibitem[{Yasarla} et~al.(2020){Yasarla}, {Perazzi}, and
  {Patel}]{ryasarla_UMSN}
R.~{Yasarla}, F.~{Perazzi}, and V.~M. {Patel}.
\newblock Deblurring face images using uncertainty guided multi-stream semantic
  networks.
\newblock \emph{IEEE Transactions on Image Processing}, 29, 2020.

\bibitem[Zhang et~al.(2018)Zhang, Chen, Wang, Hu, Zuo, and
  Hancock]{zhang2018face}
Zhihong Zhang, Xu~Chen, Beizhan Wang, Guosheng Hu, Wangmeng Zuo, and Edwin~R
  Hancock.
\newblock Face frontalization using an appearance-flow-based convolutional
  neural network.
\newblock \emph{IEEE Transactions on Image Processing}, 28\penalty0 (5), 2018.

\bibitem[Zhao et~al.(2017)Zhao, Xiong, Jayashree, Li, Zhao, Wang, Pranata,
  Shen, Yan, and Feng]{zhao2017dual}
Jian Zhao, Lin Xiong, Panasonic~Karlekar Jayashree, Jianshu Li, Fang Zhao,
  Zhecan Wang, Panasonic~Sugiri Pranata, Panasonic~Shengmei Shen, Shuicheng
  Yan, and Jiashi Feng.
\newblock Dual-agent gans for photorealistic and identity preserving profile
  face synthesis.
\newblock In \emph{Advances in neural information processing systems}, 2017.

\bibitem[Zhou et~al.(2020)Zhou, Liu, Liu, Liu, and Wang]{Zhou_2020_CVPR}
Hang Zhou, Jihao Liu, Ziwei Liu, Yu~Liu, and Xiaogang Wang.
\newblock Rotate-and-render: Unsupervised photorealistic face rotation from
  single-view images.
\newblock In \emph{IEEE/CVF Conference on Computer Vision and Pattern
  Recognition (CVPR)}, June 2020.

\bibitem[Zhu et~al.(2020)Zhu, Wu, Chen, Vesdapunt, and Wang]{Zhu_2020_CVPR}
Wenbin Zhu, HsiangTao Wu, Zeyu Chen, Noranart Vesdapunt, and Baoyuan Wang.
\newblock Reda:reinforced differentiable attribute for 3d face reconstruction.
\newblock In \emph{CVPR}, 2020.

\end{thebibliography}
\end{document}


\maketitle

\section{Additional Implementation Details}
This section clarifies some details regarding the losses in the training objective of viewpoint encoder $E_{3D}$ (section 3.4).

\subsection{Training Objectives}

\noindent \textbf{Reconstruction Losses for $E_{3D}$.}
As mentioned in section 3.4 of the paper, the training objective for the viewpoint encoder $E_{3D}$ is given by
\begin{equation}
\min_{E_{3D}}\sum_{i=1}^{n} \sum_{\nu=1}^{\nu_{i}} \ell_{im} \big( \phi \left(E_{3D}(x_i^\nu) \right), y_i^\nu \big) + \ell_{3D} \big( E_{3D}(x_i^\nu), y_i^\nu \big) + \lambda_c ( |\alpha_i|^2+ |\beta_i|^2 + |\delta_i|^2 ),
\end{equation}
\noindent where $\ell_{im}$ and $\ell_{3D}$ represents the following combination of different reconstruction losses:
\begin{align*}
\ell_{im}\Big( x, y \Big)&=\lambda_{id}\mathcal{L}_{id}\Big( x, y \Big) + \lambda_{edge}\mathcal{L}_{edge}\Big( x, y \Big) + \lambda_{data}\big|x-y\big|\\
\ell_{3D}\Big( x, y \Big)&=\lambda_{lan}\mathcal{L}_{lan}\Big( x, y \Big) + \lambda_{mview}\mathcal{L}_{mview}\Big( x \Big)\\
\mathcal{L}_{lan}\Big( x, y \Big)&=\big|\mathcal{Q}_{basel}(x) - \mathcal{Q}_{image}(y) \big|^{2}_{2}\\
\mathcal{L}_{mview}\Big( x \Big)&=\sum_{\nu}|\Bar{\alpha} - \alpha^{v}|^2+ |\Bar{\beta} - \beta^{v}|^2 + |\Bar{\delta} - \delta^{v}|^2.
\end{align*}
\noindent where, $\lambda_{data}=5$, $\lambda_{id}=0.5$, $\lambda_{edge}=30$, $\lambda_{mview}=0.25$ and $\lambda_{lan}=1$. 
We use the perspective camera model in the renderer $\phi$, with an empirically selected focal length for the $3 \mathrm{D}$-$2 \mathrm{D}$ projection.
The term $\mathcal{L}_{lan}$ is a MSE between 2D projections of facial landmarks of the predicted mesh and pre-computed landmarks in sharp images.
$\mathcal{Q}_{basel}$ projects the 3D landmark vertices of the reconstructed mesh onto the image (obtaining 68 facial landmarks), and $\mathcal{Q}_{image}$ extracts landmarks using the method of \cite{bulat2017far} from the ground-truth targets. 
$\mathcal{L}_{mview}$ ensures that the identity, texture, and expression parameters of the BFM are consistent across views for samples of our multi-view dataset.


\section{Additional Qualitative Results}

\begin{figure*}[!t]
\centering
\includegraphics[width=1.0\textwidth]{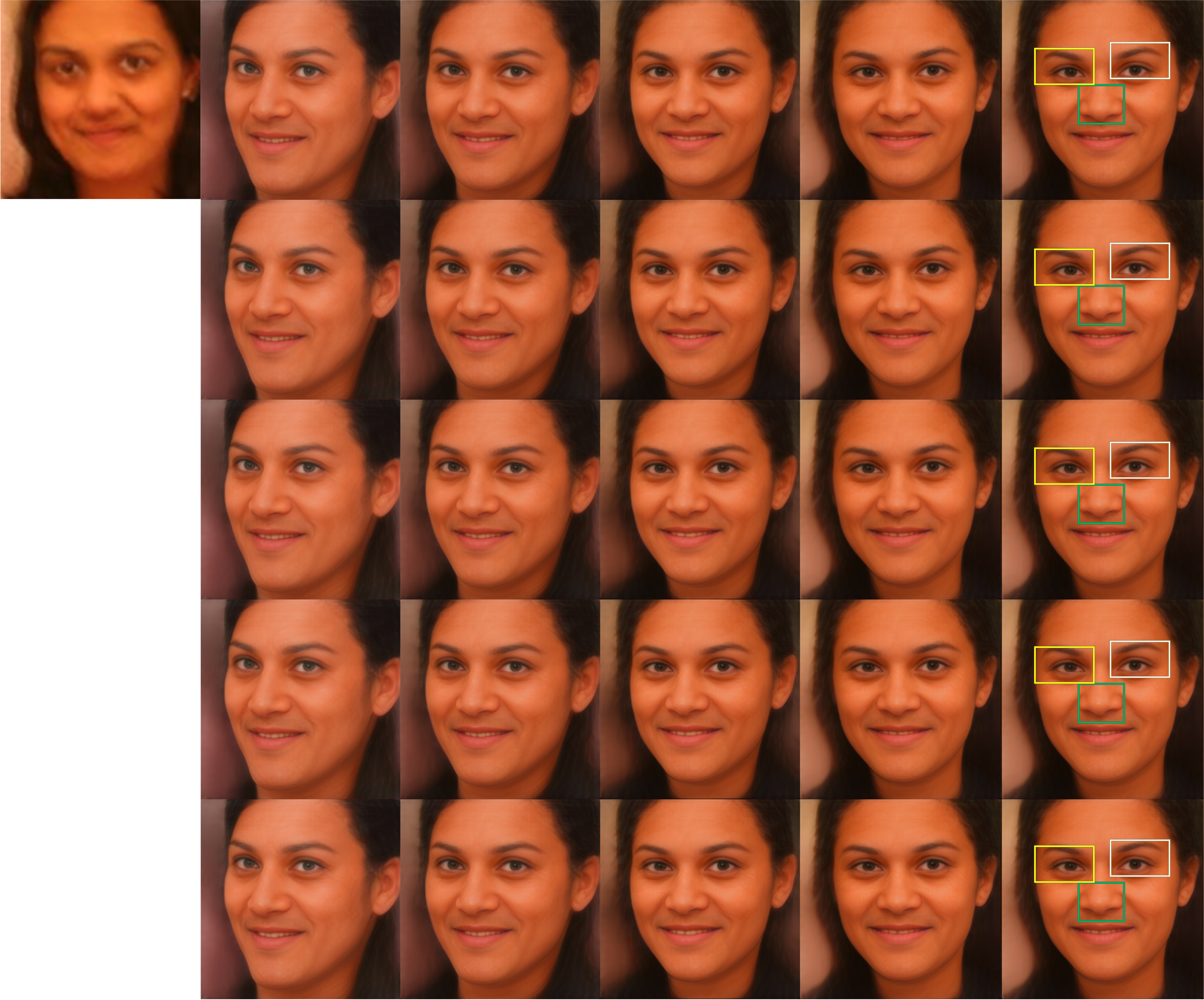}
\caption{
\textbf{Qualitative sample on real-world motion blurred face.}
The first column corresponds to the blurry input image. All the other columns are output sequences rotated by a different amount. Rows from 1 to 5 correspond to the appropriate frame in the output sequence. The last column is the copy of the previous one with rectangles on top of different facial regions. Rectangles are at a fixed location with respect to the image in all frames. Note how the both eyes and the nose move upwards 
as we go from the top to the bottom.}
\label{fig:qualitative_real}
\end{figure*}
\begin{figure*}[!t]
\centering
\includegraphics[width=1.0\textwidth]{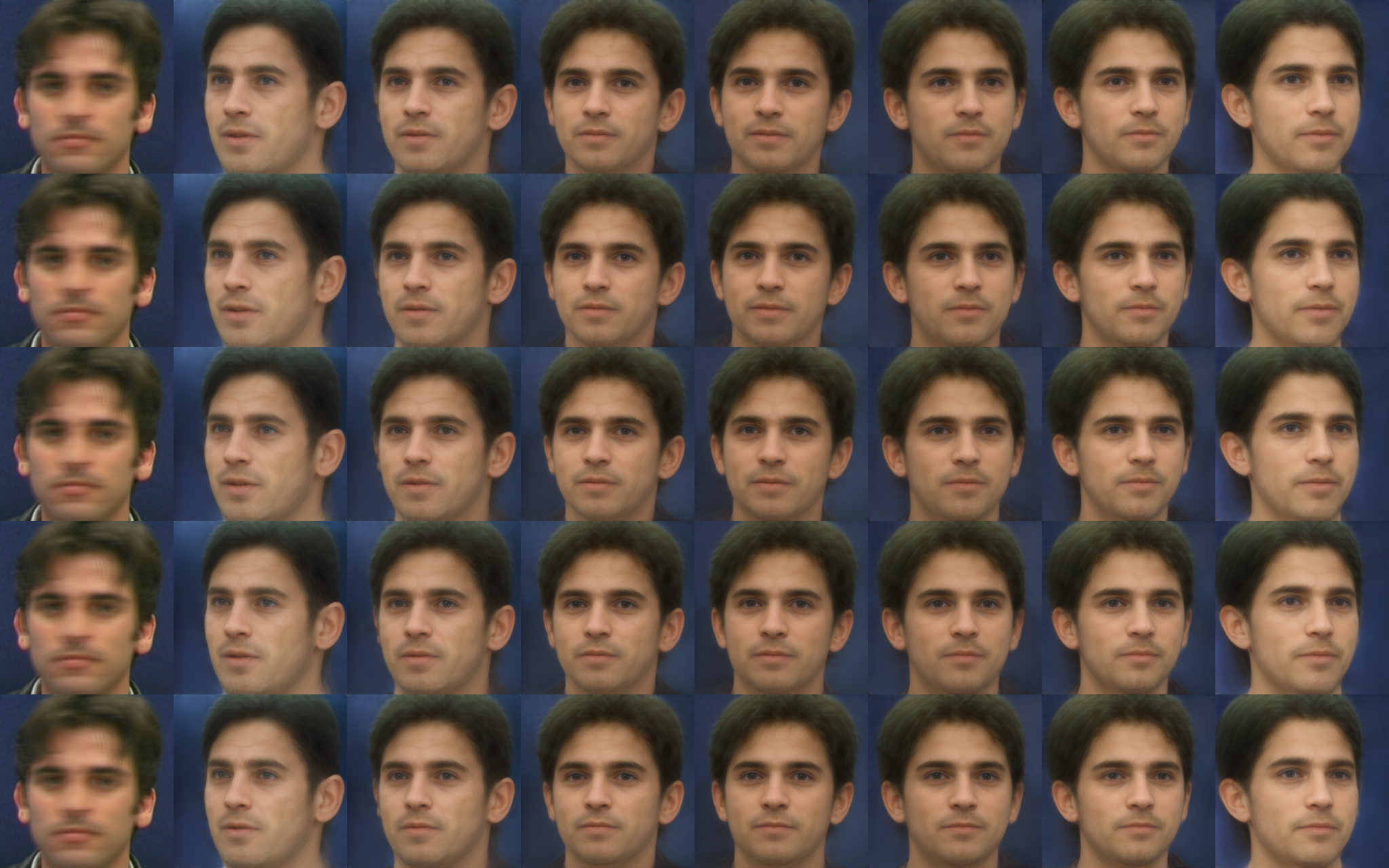}
\includegraphics[width=1.0\textwidth]{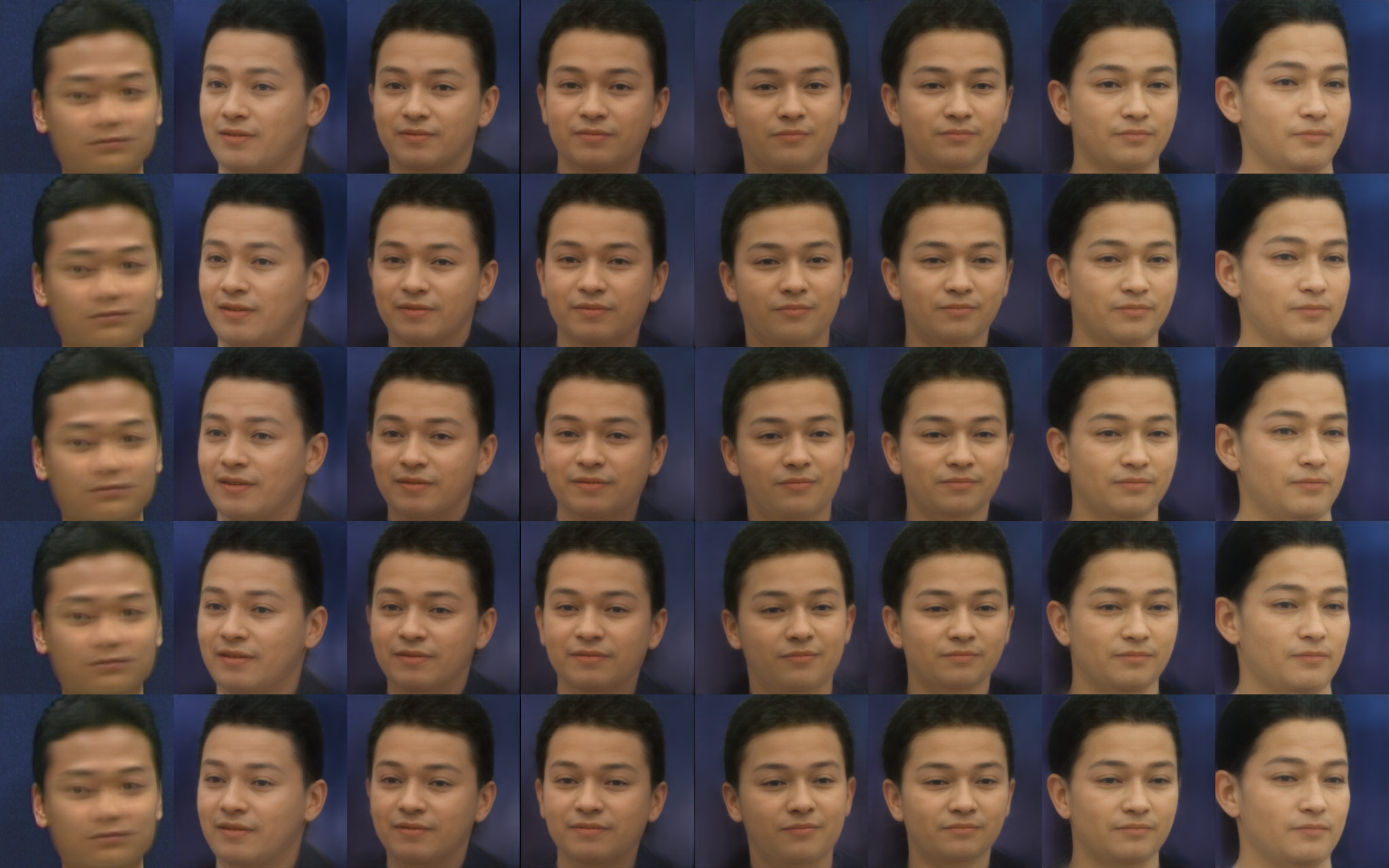}
\caption{
\textbf{Qualitative samples on VIDTIMIT.}
The first column corresponds to the blurry input image. All the other columns are output sequences rotated by a different amount. Rows from 1 to 5 correspond to the appropriate frame in the output sequence.}
\label{fig:qualitative_1}
\end{figure*}
\begin{figure*}[!t]
\centering
\includegraphics[width=1.0\textwidth]{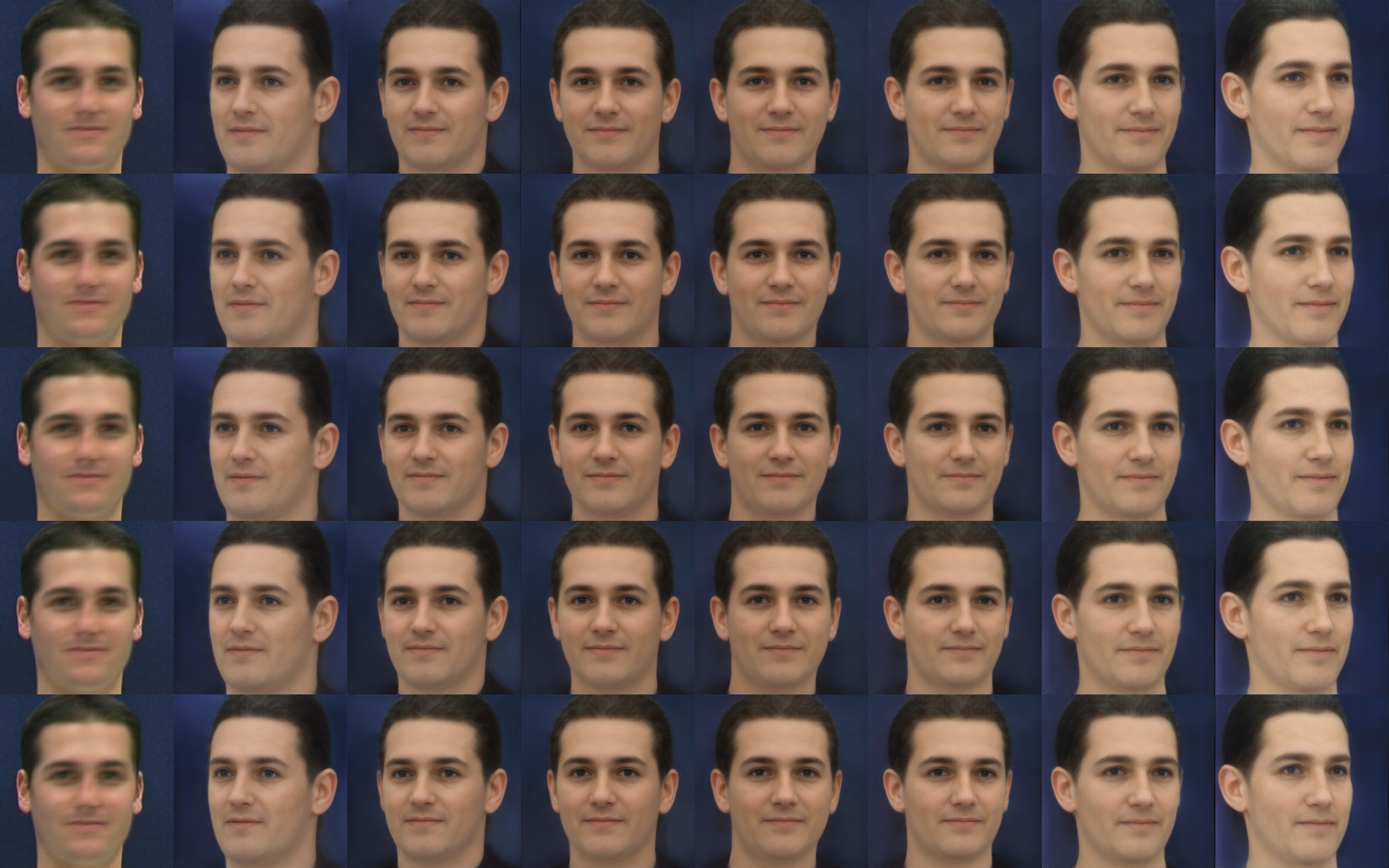}
\includegraphics[width=1.0\textwidth]{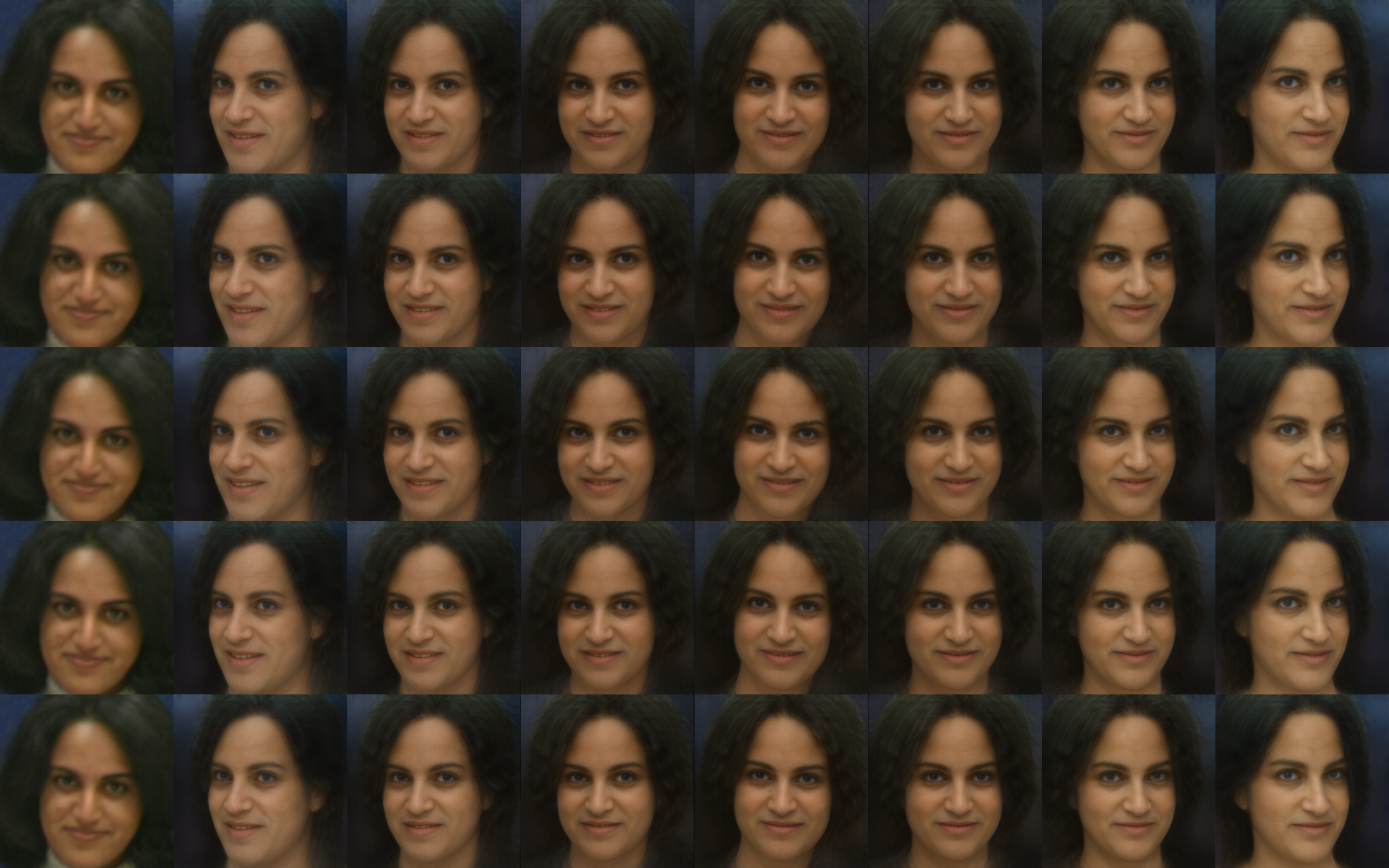}
\caption{
\textbf{Qualitative samples on VIDTIMIT.}
The first column corresponds to the blurry input image. All the other columns are output sequences rotated by a different amount. Rows from 1 to 5 correspond to the appropriate frame in the output sequence.}
\label{fig:qualitative_2}
\end{figure*}

A real-world deblurring example is presented in Figure \ref{fig:qualitative_real}. Some more qualitative examples of our multi-view reconstructions  on VIDTIMIT\cite{VIDTIMIT} can be found in Figures \ref{fig:qualitative_1} and \ref{fig:qualitative_2}.

\clearpage
\bibliography{egbib}